\documentclass[lettersize,journal, ]{IEEEtran}
\usepackage{amsmath,amsfonts}
\usepackage{algorithmic}
\usepackage{array}
\usepackage{textcomp}
\usepackage{stfloats}
\usepackage{url}
\usepackage{verbatim}
\usepackage{graphicx}
\def\BibTeX{{  B\kern-.05em{\sc i\kern-.025em b}\kern-.08em
    T\kern-.1667em\lower.7ex\hbox{E}\kern-.125emX}}
\usepackage{balance}
\usepackage{verbatim}
\usepackage{amsfonts}
\usepackage{amssymb}
\usepackage{stfloats}
\usepackage{cite}
\usepackage{soul}
\usepackage{setspace}
\usepackage{graphicx}
\usepackage{psfrag}
\usepackage{amsmath}
\usepackage{array}
\usepackage{epstopdf}
\usepackage{authblk}
\usepackage{graphicx} 
\usepackage{amsthm} 
\usepackage{lipsum}
\usepackage{verbatim} 
\usepackage{authblk}
\usepackage{mathtools}
\usepackage{cuted}
\usepackage[lined,boxed,ruled]{algorithm2e}
\usepackage{booktabs}
\usepackage{subfigure}

\usepackage{hyperref}
\hypersetup{colorlinks=false,
            linkcolor=red,
            anchorcolor=green,
            citecolor=black}

\usepackage[usenames]{color}

\newtheorem{Lemma}{Lemma}
\newtheorem{Theorem}{Theorem}

\newtheorem{Assumption}{Assumption}

\newtheorem{Remark}{Remark}

\newtheorem{Corollary}{Corollary}
\begin{document}

\title{Optimal Batch-Size Control for Low-Latency Federated Learning with Device Heterogeneity}

\author{{  {Huiling~Yang}, {Zhanwei~Wang}, and {Kaibin~Huang}                   }

\thanks{H.~Yang, Z.~Wang, and K. Huang are with the Department of Electrical and Electronic Engineering, 
The University of Hong Kong, Hong Kong SAR, China (Email: \{hlyang, zhanweiw,  huangkb\}@eee.hku.hk).
 Corresponding authors: Z. Wang; K. Huang.
}
}
\maketitle

\begin{abstract}

\emph{Federated learning} (FL) has emerged as a popular approach for collaborative machine learning in \emph{sixth-generation} (6G) networks, primarily due to its privacy-preserving capabilities. The deployment of FL algorithms is expected to empower a wide range of \emph{Internet-of-Things} (IoT) applications, e.g., autonomous driving, augmented reality, and healthcare. The mission-critical and time-sensitive nature of these applications necessitates the design of low-latency FL frameworks that guarantee high learning performance. In practice, achieving low-latency FL faces two challenges: the overhead of computing and transmitting high-dimensional model updates, and the heterogeneity in \emph{communication-and-computation} (C$^2$) capabilities across devices. To address these challenges, we propose a novel C$^2$-aware framework for optimal batch-size control that minimizes \emph{end-to-end} (E2E) learning latency while ensuring convergence. The framework is designed to balance a fundamental C$^2$ tradeoff as revealed through convergence analysis. Specifically, increasing batch sizes improves the accuracy of gradient estimation in FL and thus reduces the number of communication rounds required for convergence, but results in higher per-round latency, and vice versa. The associated problem of latency minimization is intractable; however, we solve it by designing an accurate and tractable surrogate for convergence speed, with parameters fitted to real data. This approach yields two batch-size control strategies tailored to scenarios with slow and fast fading, while also accommodating device heterogeneity. Extensive experiments using real datasets demonstrate that the proposed strategies outperform conventional batch-size adaptation schemes that do not consider the C$^2$ tradeoff or device heterogeneity.


\end{abstract}

\section{Introduction}

A promising new usage scenario for \emph{sixth-generation} (6G) mobile networks is edge learning, which refers to the ubiquitous deployment of machine learning algorithms at the network edge to distill \emph{Artificial Intelligence} (AI) from distributed data~\cite{saad2019vision,liu2025integrated,chen2025space}. This emerging platform aims to support a wide range of intelligent mobile applications, from personal assistants to autonomous vehicles. \emph{Federated Learning} (FL) has become the mainstream framework for edge learning due to its privacy-preserving nature, enabling collaborative model training among devices without the need to transmit raw data~\cite{mcmahan2017communication, yang2019federated, kairouz2021advances}. As 6G aims to connect high-mobility devices (e.g., vehicles and \emph{unmanned aerial vehicles} (UAVs)), there is an increasing need for low-latency FL implementations. Specifically, these devices can participate in FL only for brief periods before moving out of an edge server’s coverage area, which can lead to stale model updates and degraded learning performance~\cite{wu2020fedhome,parekh2023gefl}. Moreover, accelerating the FL process helps efficient utilization of \emph{communication-and-computation} (C$^2$) resources. However, realizing low-latency FL in mobile networks presents two key challenges. First, given the iterative nature of FL algorithms, 
computing and transmitting high-dimensional model updates over multiple rounds creates both computation and communication bottlenecks.
Second, the heterogeneity in C$^2$ capabilities across participating devices can slow down model convergence. To address these challenges, we propose a novel batch-size control framework that optimally balances C$^2$ operations, with the goal of minimizing \emph{end-to-end} (E2E) learning latency while maintaining robust satisfactory learning performance. 

There are several approaches to overcoming the aforementioned communication bottleneck in FL systems. One popular approach is source compression to reduce communication overhead~\cite{konevcny2016federated,jiang2022model,lin2021deploying,shlezinger2020uveqfed,qiao2021communication}. Relevant techniques include model pruning to remove redundant weights~\cite{jiang2022model,lin2021deploying}, low-resolution quantization~\cite{shlezinger2020uveqfed}, and low-rank factorization, which approximates large weight matrices as products of sparse factors~\cite{qiao2021communication}. A second approach is split learning involving partitioning the global model into two components for deployment on devices and the server, respectively. This strategy reduces communication overhead by requiring devices to transmit only partial gradients and intermediate activations~\cite{thapa2022splitfed}. The third approach is FL with \emph{Over-the-Air Computing} (AirComp)~\cite{zhu2019broadband,  yang2020federated}. This class of techniques enables simultaneous access by realizing over-the-air aggregation of model updates when transmitted concurrently by multiple devices. However, due to the use of uncoded linear analog modulation, the techniques are vulnerable to channel noise and interference~\cite{wang2024spectrum,wang2025airbreath}, and also encounter the issue of compatibility with existing digital communication infrastructures~\cite{amiri2020federated}.

Besides the communication bottleneck, a computation bottleneck also significantly contributes to the E2E latency of FL systems as driven by two primary factors. 
First, the increasing complexity of AI models (e.g., large language models) and the growing scale of datasets, combined with the limited energy and processing capabilities of edge devices, exacerbate mobile computing latency~\cite{goyal2017accurate}. 
This latency is comparable to communication latency and should not be overlooked. 
For instance, executing a single FL iteration of ResNet-50 (with approximately 26 million parameters) on a GPU such as the NVIDIA GTX 1080 Ti can take 400–600 ms, which is on par with the 800 ms required to transmit the same model over a 1 Gbps wireless link~\cite{shi2022talk}. 
Second, device heterogeneity in C$^2$ capabilities further slows down the learning process. Variations in computation speed (e.g., from low-power ARM Cortex-M7~\cite{armCortexM7} to high-performance Apple A18 Pro~\cite{appleA18Pro}) and channel conditions among edge devices lengthen server-side waiting times during aggregation due to slower devices, commonly referred to as "stragglers".
The literature addresses the straggler issue primarily through two types of strategies. 
The first employs asynchronous aggregation protocols, enabling edge devices to independently and concurrently update the global model without waiting for other devices~\cite{xie2019asynchronous,wang2022asynchronous,xu2023asynchronous}. 
While these protocols effectively reduce per-round latency, they are susceptible to stale updates, which can degrade overall convergence performance~\cite{mcmahan2017communication,chen2016revisiting}. 
The second strategy maintains synchronous model updates while incorporating system-level controls to mitigate the impact of stragglers, such as bandwidth allocation~\cite{xu2021bandwidth,xu2020client}, transmit power control~\cite{shi2020joint,zeng2021wirelessly}, and batch-size adaptation~\cite{ma2021adaptive,park2022amble}. 
Among these, batch-size control offers a straightforward, globally coordinated mechanism to simultaneously regulate device updating latency
and the learning performance. 
Traditionally, this approach mitigates computational heterogeneity by increasing batch sizes for faster devices and decreasing them for slower ones~\cite{ma2021adaptive,park2022amble}.
However, prior work has primarily targeted computer-cluster implementation, focusing solely on the computation aspect without considering the effects of wireless channels—an essential factor in wireless FL systems.

Recent research on batch-size optimization in FL systems has progressed beyond simply mitigating stragglers to jointly addressing heterogeneity in both computation and communication~\cite{song2024optimal,ren2020accelerating,liu2020adaptive,liu2023dynamite}. 
Several studies have examined this challenge under the constraint of a fixed global batch size—the total number of samples processed per round across all devices—with the objective of minimizing per-round latency~\cite{song2024optimal,ren2020accelerating}. 
These works often make restrictive assumptions or consider specific settings, such as particular multi-access protocols~\cite{song2024optimal} or the strong convexity of the loss function~\cite{ren2020accelerating}.
Subsequent research has introduced more sophisticated controls, 
including co-optimizing local batch size with model compression ratios~\cite{liu2020adaptive} and the joint tuning of batch size and aggregation frequency~\cite{liu2023dynamite}, to further enhance learning performance. 
In view of prior work, their common limitation lies in the assumption of a fixed number of training rounds, which fails to account for the effects of batch size on convergence and the resulting fundamental C$^2$ tradeoff, as explained below.
On one hand, a larger batch size improves the accuracy of gradient estimation and can reduce the number of communication rounds needed for convergence. 
On the other hand, increasing the batch size also raises the per-round 
latency due to greater
computation overhead. 
Then effectively controlling the batch size to balance this tradeoff presents a viable direction for improving FL performance, but it has received limited attention thus far.
A tangentially related work proposes exponentially increasing batch sizes across rounds. 
This approach is obviously suboptimal and furthermore relies on simplifying assumptions such as static channels and homogeneous device capabilities~\cite{shi2022talk}. 
The dynamic control of batch sizes with adaptation to device heterogeneity and fluctuating channel states remains a largely open area. It is discovered in this work to be a promising direction for accelerating FL.

In this paper, we present a novel framework of batch-size control to enable low-latency FL systems with device heterogeneity. 
The design objective is to jointly optimize both the global batch size and local batch sizes at devices to minimize the E2E learning latency under a constraint on learning accuracy. 
The key contributions and findings of this work are summarized as follows.

\begin{itemize}
    \item \textbf{Theoretic characterization of C$^2$ tradeoff:} To this end, convergence analysis is conducted to reveal the aforementioned fundamental C$^2$ tradeoff. Specifically, we theoretically derive an upper bound on the convergence rate by leveraging weighted-sum gradient aggregation and the inverse proportional relationship between batch size and gradient estimation error. From the analysis, the convergence speed in round is found to be inversely proportional to the global batch size. On the other hand, increasing the batch size lengthens the per-round latency. Combining these findings quantifies the E2E latency in seconds and thus characterizes the C$^2$ tradeoff. Balancing this tradeoff to reduce the latency necessitates the optimization of batch size in the sequel.

    \item \textbf{Optimal batch-size control with slow fading:} In this case, channel states are assumed to be heterogeneous across users but remain constant throughout the learning duration. An E2E latency minimization problem is formulated to jointly optimize the global batch size, convergence speed, and local batch-size allocation. However, this results in an intractable Mixed-Integer Nonlinear Programming (MINLP) problem. To address this challenge, we develop a tractable approach that accurately approximates the convergence result using a parameterized function, with parameters fitted to real data.
    The approximate MINLP problem is then solved in two steps: first, we derive the optimal batch-size allocation across devices for a given global batch size; second, conditioned on this allocation strategy, the optimal global size is obtained in closed form.
    The resulting solution reveals that each optimal local batch size scales \emph{linearly} with the associated device's computational speed and \emph{inversely} with its communication latency.

    \item \textbf{Adaptive batch-size control with fast fading:}
    Building on the preceding solution, we design an algorithm for adaptive batch-size control in the presence of fast-fading channels that vary across rounds. 
    The algorithm begins by initializing the global and per-device batch sizes optimized based on long-term channel statistics. 
    During the learning process, these batch sizes are dynamically adapted to the channel states as detailed below.
    Essentially, the algorithm seeks to equalize latency across devices—a strategy proven to be optimal—by allocating larger batch sizes to devices with higher computational speeds.      
    Consequently, the accuracy of collaborative gradient estimation is improved to accelerate convergence without increasing per-round latency, thereby reducing the total training time of the FL process.

    \item \textbf{Experimental Results:}
     Experiments on real-world datasets are conducted to evaluate the performance of the proposed algorithms. Compared with conventional designs without awareness of C$^2$ tradeoff or device heterogeneity, empirical results demonstrate that the proposed algorithms consistently achieve comparable learning performance while achieving substantial reductions in E2E learning latency.

\end{itemize}

The remainder of the paper is organized as follows. Section~\ref{sec:system_models} introduces the system model and performance metrics. The convergence analysis is presented in Section~\ref{sec:convergence_analysis}. Section~\ref{sec:fixed_snr_optimization} derives the optimal batch-size control with slow fading, and the framework is extended to adaptive batch-size control with fast fading in Section~\ref{sec:Adaptive_Batch_Size_Control}. Experimental results are discussed in Section~\ref{sec:experimental results}, and conclusions are drawn in Section~\ref{sec:conclusion}.

\section{Models and  Metrics}

\label{sec:system_models}
\begin{figure*}[t]
    \centering
    \includegraphics[width=16cm]{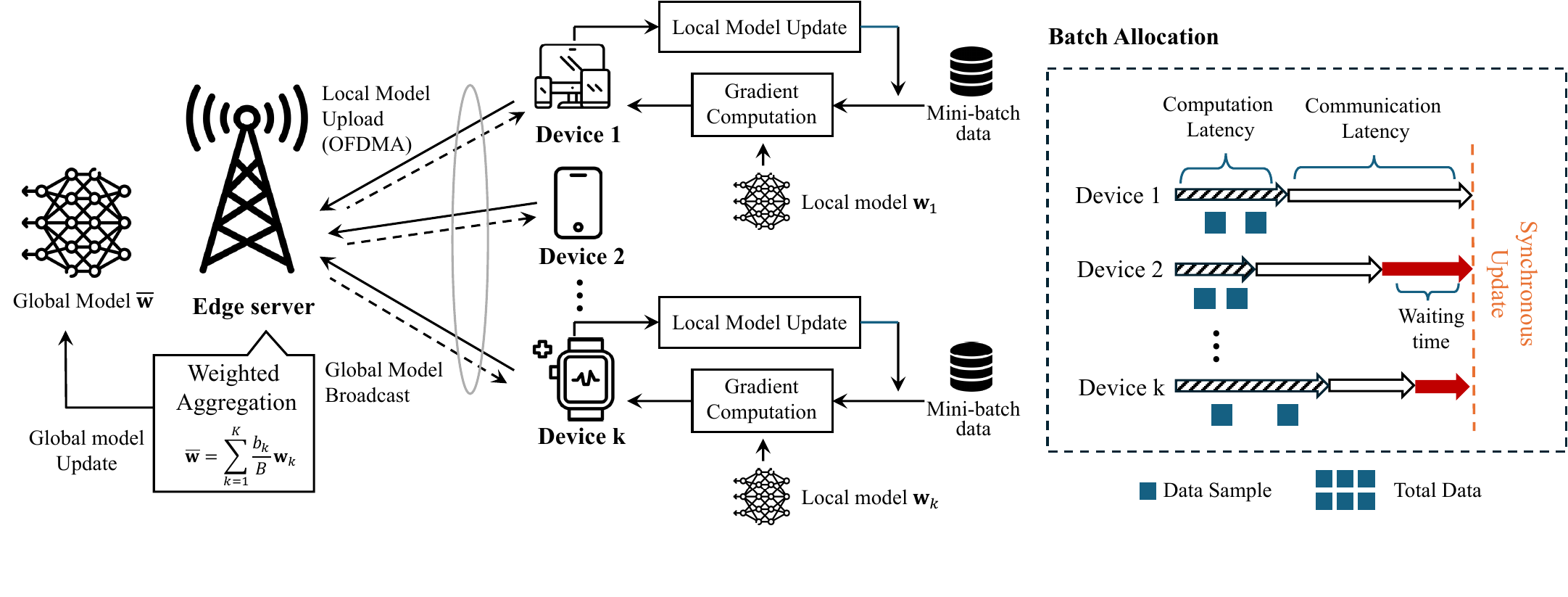} 
    \vspace{-5mm}
    \caption{FL system with synchronous model updates.}
    \label{fig:system_model}
    \vspace{-3mm}
\end{figure*}

As illustrated in Fig.~\ref{fig:system_model}, the considered FL system comprises a single server and 
$K$ distributed devices.
The server coordinates these devices to collaboratively train an AI model using a distributed machine learning algorithm.
The system model and performance metrics are detailed in the following subsections.

\subsection{Federated Learning Model}
\label{subsec:Federated_Learning_Model}

In the FL system, the edge server schedules the $K$ devices to iteratively optimize the global model weight  \( \overline{\mathbf{w}} \in \mathbb{R}^D  \) using the local datasets.
Let \( \mathcal{D}_k \) denote the dataset of device \( k \), comprising \(|\mathcal{D}_k|\) sample-label pairs \(\{(\mathbf{x}_j, y_j)\}\in \mathcal{D}_k \), where $\mathbf{x}_j$ is a data sample and $y_j$ is the corresponding label. 
The learning process pursues the minimization of an empirical global loss function defined as
\begin{equation}
     F(\overline{\mathbf{w}}) = \frac{1}{\sum_{k=1}^K|\mathcal{D}_k|} \sum_{k=1}^K \sum_{(\mathbf{x}_j,y_j) \in \mathcal{D}_k} l(\overline{\mathbf{w}}; (\mathbf{x}_j, y_j)),
    \label{eq:global_objective}
\end{equation}
where  \( l(\overline{\mathbf{w}}; (\mathbf{x}_j, y_j)) \) is the sample-wise loss function that quantifies the prediction error between the model output based on \(\mathbf{x}_j\) and the true label \( y_j \).

We consider a synchronous FL system where the iterative model update proceeds over multiple communication rounds, indexed by $n \in \{1, 2, \dots\}$, until a satisfactory performance level is achieved. In each communication round $n$, the global model is updated through two steps.
In the first step, each device updates its local model weights based on the current global model and its local dataset.
Specifically, at the beginning of round $n$, the server broadcasts the current global model, denoted as $\overline{\mathbf{w}}_n$, to all participating devices.
Upon reception,  device $k$ initializes its local model as $\mathbf{w}_{k,n}^{(0)}=\overline{\mathbf{w}}_n$
and subsequently executes $H$ local iterations of \emph{stochastic gradient descent} (SGD) using its local dataset $\mathcal{D}_k$.
For the $t$-th local iteration ($1\le t \le H$),  device $k$ samples a mini-batch $\mathcal{B}_{k,n}^{(t)}\subset\mathcal{D}_k$ with the size of $b_{k,n}=|\mathcal{B}_{k,n}^{(t)}|$ and computes the gradient as
\begin{equation}
    \mathbf{g}_{k,n}^{(t)}  = \frac{1}{b_{k,n}} \sum_{(\mathbf{x}_j, y_j) \in \mathcal{B}_{k,n}^{(t)}} \nabla l(\mathbf{w}_{k,n}^{(t)}; (\mathbf{x}_j, y_j)),
    \label{eq:stoch_grad}
\end{equation}
where $\nabla$ represents the gradient operation.
Then SGD is performed to update the local model by
\begin{equation}
\mathbf{w}_{k,n}^{(t)} = \mathbf{w}_{k,n}^{(t-1)} - \eta \cdot \mathbf{g}_{k,n}^{(t-1)},
\label{eq:local_update_clean}
\end{equation}
where $\eta$ denotes the learning rate. After completing $H$ local updates, device $k$ uploads the final local model $\mathbf{w}_{k,n}^{(H)}$ to the server.

In the second step, the server updates the global model by aggregating the locally updated models from all devices via a weighted averaging scheme
\begin{equation}
    \overline{\mathbf{w}}_{n+1} = \sum_{k=1}^K \frac{b_{k,n}}{B_n} \mathbf{w}_{k,n}^{(H)},
    \label{eq:aggregation}
\end{equation}
where \( B_n = \sum_{k=1}^K b_{k,n} \) represents the global batch size across all devices in round \( n \). 
This weighted aggregation accounts for the effect of gradient quality, determined by varying batch sizes, on the model update, a technique widely adopted in related works~\cite{song2024optimal,ren2020accelerating}.
Gradients computed using larger batch sizes are considered to have higher quality and exert a greater effect on the global model update.



\subsection{Device Heterogeneity Model}

We consider device heterogeneity in both computation and communication, which leads to varying latencies in computing and uploading local updates. These aspects are discussed in detail below.

\subsubsection{Computation Latency} 
\label{subsec:computation_model}

Given the limited processing resources at device $k$, the computation of local updates incurs a latency denoted as \( T_{k,n}^{\sf cmp} \).
Let $W$ represent the per-sample computational workload, measured in \emph{floating-point operations} (FLOPs).
The computation latency of device $k$ at round $n$ is given by
\begin{equation}
    T_{k,n}^{\sf cmp}  = \frac{HWb_{k,n}}{f_k},
    \label{eq:computation_time}
\end{equation}
where $H$ denotes the number of local updates per round, \( b_{k,n} \) is the allocated batch size for device \( k \) in the current round $n$ and \( f_k \) (in FLOPs/s) represents the computation speed of device \( k \), quantifying the computation heterogeneity over devices.
Note that the computation latency scales linearly with the batch size \( b_{k,n} \), establishing batch size as a control variable for latency optimization in FL systems~\cite{zeng2021wirelessly}.

\subsubsection{Communication Latency}
\label{subsec:communication_model}

For model uploading from distributed devices to the edge server, we consider an \emph{Orthogonal Frequency Division Multiple Access} (OFDMA) scheme with $K$ orthogonal sub-bands, each having a bandwidth of $B_W$ Hz. 
Consider the communication round $n$.
The transmission rate for device \( k \) in round \( n \) is given by
\begin{equation}
    R_{k,n} = B_W \log_2\left(1 + \frac{P_{k}|h_{k,n}|^2}{B_WN_0}\right) ,
    \label{eq:upload_rate}
\end{equation}
where \( P_{k} \) represents the transmit power of device \( k \), \( N_0 \) denotes the power spectral density of the \emph{additive white Gaussian noise} (AWGN), and \( h_{k,n} \) is the \emph{independent and identically distributed} (i.i.d.) channel coefficient for the link between device \( k \) and the server in round \( n \), and is assumed to be known at both the transmitter and receiver.
We consider two types of fading channel models for batch-size optimization. One setting is slow-fading channels, in which the channel coefficients remain constant across rounds throughout the learning process, i.e., $h_{k,n}=h_k, \forall n$. The other pertains to fast‑fading channels, where $h_{k,n}$ varies with round index $n$. 

For the model uploading of device $k$, its model parameter $\mathbf{w}_k\in\mathbb{R}^D$ is quantized using a sufficiently large number of bits $Q$, rendering the quantization error negligible.
Consequently, the uplink payload size is $
q = D \times Q $ bits.
The resulting uplink transmission latency for device \( k \) in round \( n \) is thus given by
\begin{equation}
 T^{\sf{cmm}}_{k,n} = \frac{q}{R_{k,n}}.
    \label{eq:upload_time}
\end{equation}
Note that $T^{\sf{cmm}}_{k,n}$ remains constant under slow fading and varies i.i.d across rounds under fast fading.
The latencies resulting from model broadcasting and aggregation are assumed to be negligible due to the server's sufficient transmit power and processing capability, and are therefore omitted from the analysis.

\subsection{Performance Metrics}

Two metrics for measuring learning performance are considered and defined as follows.

\subsubsection{Validation Accuracy \& Convergence Rate}
The learning accuracy is experimentally measured by the validation accuracy during the learning process on the real dataset.
For the  tractable analysis, the learning performance for non‑convex optimization problems is generally assessed by the convergence rate, denoted as $\mathcal{R}_{\sf con}$, which is measured by the expected average gradient norm~\cite{zeng2021wirelessly,shi2022talk,zhou2017convergence}
\begin{equation}
\label{eq:R_con}
   \mathcal{R}_{\sf con}= \frac{1}{N}\sum_{n=0}^{N-1} \mathbb{E}\left[\|\nabla F(\overline{\mathbf{w}}_n)\|^2\right],
\end{equation}
where $\mathbb{E}\left[\|\nabla F(\overline{\mathbf{w}}_n)\|^2\right]$ denotes the expected gradient norm of model weight \( \overline{\mathbf{w}}_n \) at round \(n\), and \( N \) is the total number of communication rounds.
This metric reflects the optimization progress in non-convex loss landscapes, where monitoring the gradient norm is crucial for evaluating proximity to stationary points. A smaller gradient norm suggests that the model is approaching a local minimum or saddle point, indicating potential convergence.

\subsubsection{E2E Learning Latency}  
In synchronous FL systems, E2E latency characterizes the wall-clock time to achieve successful convergence.
Consider the successful convergence indicated by $\mathcal{R}_{\sf con}\leq \epsilon$ where $\epsilon$ defines the required learning threshold.
Let \(N_\epsilon\) denote the associated minimum number of communication rounds to achieve an \(\epsilon\)-level global convergence accuracy, termed as convergence round.
The  E2E learning latency with $\epsilon$-guaranteed convergence, denoted as \(T_{\sf e2e}\), is governed by the bottleneck device in each round, which is given by
\begin{equation}
\begin{aligned}
  T_{\sf e2e} &= \sum_{n=1}^{N_{\epsilon}}\tau_n,
    \label{eq:total_latency}
\end{aligned}
\end{equation}  
where $\tau_n$ represents the per-round latency, given by
\begin{equation}
\label{eq:per_round_latency}
    \tau_n=\max_{k} \left(T^{\sf cmp}_{k,n} + T^{\sf cmm}_{k,n}\right).
\end{equation}
Here, \(T^{\sf cmp}_{k,n}\) and \(T^{\sf cmm}_{k,n}\) denote the computation and communication latencies for device \(k\) in round \(n\), respectively.
Note that per-round latency is determined by the straggler device with the slowest combined latency.

\section{Analysis of Batch-size Regulated convergence}
\label{sec:convergence_analysis}
In this section, we perform a convergence analysis to quantify the effects of the number of communication rounds and the global batch size on the convergence rate.
Through this analysis, we reveal the fundamental C$^2$ tradeoff governed by the global batch size.

\subsection{Preliminaries}
\label{subsec:preliminaries}

The convergence analysis builds upon the following assumptions, which are widely adopted in non‑convex optimization for FL systems~\cite{zeng2021wirelessly,shi2022talk,zhou2017convergence,yu2019parallel,RN9}.

\begin{Assumption}[Smoothness]
\label{AS:smoothness}
The loss function $F(\cdot): \mathbb{R}^D \to \mathbb{R}$ is differentiable and satisfies the Lipschitz smoothness condition
\begin{equation}
\left\| \nabla F(\mathbf{u}) - \nabla F(\mathbf{v}) \right\| \leq L \| \mathbf{u} - \mathbf{v} \|,  \quad \forall \mathbf{u}, \mathbf{v} \in \mathbb{R}^D,
\end{equation}
where \( L > 0 \) is the Lipschitz constant.
\end{Assumption}

\begin{Assumption}[Bounded Loss Function] 
\label{AS:Bounded Loss Function}
The loss function $F(\cdot)$ is assumed to be lower bounded by
\begin{equation}
F(\mathbf{w}) \geq F^*, \quad \forall \mathbf{w} \in \mathbb{R}^D,
\end{equation} 
where $F^* \triangleq \inf_{\mathbf{w} \in \mathbb{R}^D} F(\mathbf{w})$ denotes the infimum value of the loss function.
\end{Assumption}

\begin{Assumption}[Unbiased Gradient Estimation] 
\label{AS:unbias_gradients} 
For any parameter vector $\mathbf{w} \in \mathbb{R}^D$, the   gradient computed using the local dataset is assumed to be an unbiased estimate of the true gradient, given by
\begin{equation}
    \mathbb{E}_{\mathcal{B}_k \sim \mathcal{D}_k}\left[\nabla F_k \left( \mathbf{w}; \mathcal{B}_{k} \right)\right] = \nabla F(\mathbf{w}), \quad \forall k ,
    \label{eq:unbiased_grad}
\end{equation}
where $\mathcal{B}_k$ denotes the mini-batch sampled from device $k$'s data distribution $\mathcal{D}_k$.
\end{Assumption}

\begin{Assumption}[Bounded Gradient Variance~\cite{shi2022talk}]
\label{AS:bounded_variance}  
Given the batch size of $b_{k,n}$, the variance of the estimated gradient is assumed to be upper bounded by
\begin{equation}
\mathbb{E}_{\mathcal{B}_k \sim \mathcal{D}_k} \left[ \left\| \nabla F_k\left(\mathbf{w};\mathcal{B}_k\right) - \nabla F(\mathbf{w}) \right\|^2 \right] \leq \frac{\sigma^2}{b_{k,n}}, \quad \forall k,n.
\end{equation}
\end{Assumption}

\subsection{Convergence Analysis}

Building upon the above assumptions, this subsection characterizes the convergence performance of the considered FL system in Sec.~\ref{subsec:Federated_Learning_Model}.
For analytical tractability, we fix 
the global batch size across rounds, i.e., $ B_n=B, n \in \{1, 2, \dots\}$. 
The effects of system configurations on the convergence rate are established in Theorem \ref{thm:H_avg_convergence}.

\begin{Theorem} 
\label{thm:H_avg_convergence}

Considering a fixed per-round global batch size $B$, and a learning rate $\eta$ satisfying
\begin{equation}
\eta \leq
\begin{cases}
\displaystyle \frac{1}{L}, & H = 1, \\
\displaystyle \frac{ -H + \sqrt{H(3H - 2)} }{L H(H - 1)}, & H \neq 1,
\end{cases}
\end{equation}
the convergence rate after $N$ communication rounds is upper bounded by
\begin{equation}
\begin{aligned}\label{eq:convergence_rate}
\mathcal{R}_{\sf con}
\leq & \;\underbrace{\frac{2 \left( F(\overline{\mathbf{w}}_0)  - F^*\right)}{\eta N H}}_{\text{initialization error}}\\
& + \underbrace{\frac{\eta L \sigma^2 }{B} \left(H+\frac{\eta L K (2H-1)(H-1)}{6} \right)}_{\text{batch-size-induced error}}.
\end{aligned}
\end{equation}
\end{Theorem}
\noindent\textit{Proof:} See Appendix \ref{subsec:H_avg_proof}.

The upper bound of the convergence rate is decomposed into the sum of two terms:
initialization error and batch-size-induced error. 
The former results from the gap between the initial point and the optimal model weight, decaying inversely proportional to the total number of local updates, i.e., $\mathcal{O}\left((N H)^{-1}\right)$. 
The latter scales inversely with the global batch size $B$. This reflects the reduction in gradient noise when more samples are averaged per round, thereby accelerating convergence.
Additionally, both error components depend on the number of local updates $H$.  
Increasing $H$ involves more samples per round for gradient computation and thus accelerates the decay of the initialization error, while it amplifies the batch-size-induced error due to device drift, i.e., the divergence among local models arising from data heterogeneity~\cite{yu2019parallel}.

Based on the theoretical relationship among the convergence rate, global batch size, and communication round established in Theorem~\ref{thm:H_avg_convergence}, we derive the \emph{round–batch scaling law} under a target convergence rate, as presented in Corollary~\ref{col:round_batch_constraint}.

\begin{Corollary}[Round-Batch Scaling Law]
\label{col:round_batch_constraint}
Under the convergence constraint, i.e., $\mathcal{R}_{\sf con} \leq \epsilon$, the minimum number of rounds that achieves $\epsilon$-level convergence, termed as convergence round 
$N_\epsilon$, is given by
\begin{equation}
N_{\epsilon} = \frac{\alpha}{\epsilon - \beta/B}, \quad \epsilon > \beta/B,
\label{eq:round_batch}
\end{equation}
where $\alpha = \frac{2(F(\overline{\mathbf{w}}_0) - F^*)}{\eta H}$, $\beta = \eta L \sigma^2 \left(H+\frac{\eta L K (2H-1)(H-1)}{6} \right)$ are derived from a rearrangement of \eqref{eq:convergence_rate}.
\end{Corollary}

\begin{Remark}[Communication-Computation Tradeoff]
\label{remark:Communication_Computation_Tradeoff} 
\eqref{eq:round_batch} reveals that the global batch size $B = \sum_{k=1}^K b_{k,n}$ is inversely proportional to the required number of rounds $N_\epsilon$ to attain the target convergence accuracy.
Thus, increasing $B$ reduces convergence round. However, this comes at the cost of heavier per-device computation workload in \eqref{eq:computation_time}, leading to higher per-round latency. Consequently, a fundamental tradeoff arises between more communication rounds and larger local computation overhead.
To minimize E2E latency while ensuring the desired learning performance, it is therefore crucial to balance this tradeoff. This motivates the design of optimal and adaptive batch-size control strategies in the following sections.
\end{Remark}

\section{Optimal Batch-Size Control with slow fading} 
\label{sec:fixed_snr_optimization}


In this section, we optimize batch sizes to minimize E2E latency over a slow‑fading channel while ensuring learning performance. 
Based on the estimated relation between the convergence round and global batch size, we derived the optimal batch-size strategy by balancing the C$^2$ tradeoff for the FL system with heterogeneous devices.

\label{subsec:Empirical Validation of Batch-Round Constraint and C$^2$ Tradeoff}

\subsection{Surrogate of Convergence Rate}
\label{subsec:Empirical Validation of Batch-Round Constraint}

The round–batch scaling law in Corollary~\ref{col:round_batch_constraint} describes the theoretical relationship between the convergence round \(N_\epsilon\) and the global batch size \(B\) to achieve \(\epsilon\)-level convergence. However, \(\alpha\) and \(\beta\) in \eqref{eq:round_batch} are unknown parameters that depend on the learning task, model architecture, etc. 
To tackle this challenge and enable tradeoff optimization, we consider a data-driven approach to validate the law’s form and estimate its parameters.
The parameter estimation process involves two key steps. First, data sampling is performed during offline training using a designated training dataset. Second, curve fitting is conducted to minimize the least-squares criterion between the sampled data and a predefined target formula. 

In the data sampling phase, the convergence performance is assessed by tracking a surrogate of convergence rate, i.e., the validation accuracy during the training process.
We target an accuracy threshold \(\epsilon_A\), serving as a practical proxy for the convergence threshold \(\epsilon\) in \eqref{eq:round_batch}.
For a set of feasible global batch sizes \(B^{(m)},\; m=1,2,\dots M\), we record the associated convergence round \(N_{\epsilon_A}^{(m)}\), i.e., the minimum number of rounds satisfying the target validation accuracy \(\epsilon_A\). 
To mitigate randomness, each configuration is repeated over multiple trials with independent weight initialization.

Building on the collected data samples \(\{B^{(m)},\,N_{\epsilon_A}^{(m)}\}\), we obtain the estimates of \(\alpha\) and \(\beta\) in \eqref{eq:round_batch}, denoted as $\hat{\alpha}$ and $\hat{\beta}$, respectively.
By minimizing the \emph{sum of squared residuals} (SSR)~\cite{levenberg1944method}, the estimated parameters are given as
\begin{equation}
    \hat{\alpha}, \hat{\beta}=\arg \min_{\alpha,\beta}\;\sum_{m=1}^M\left[N_{\epsilon_A}^{(m)} - \frac{\alpha}{\epsilon - \beta/B^{(m)}}\right]^2,
    \label{eq:param_est}
\end{equation}
where the optimization is subject to $\alpha > 0, \beta > 0,\epsilon > \frac{\beta}{B^{(m)}}$.

\begin{figure}[t]
    \centering
    \includegraphics[width=0.7\columnwidth]{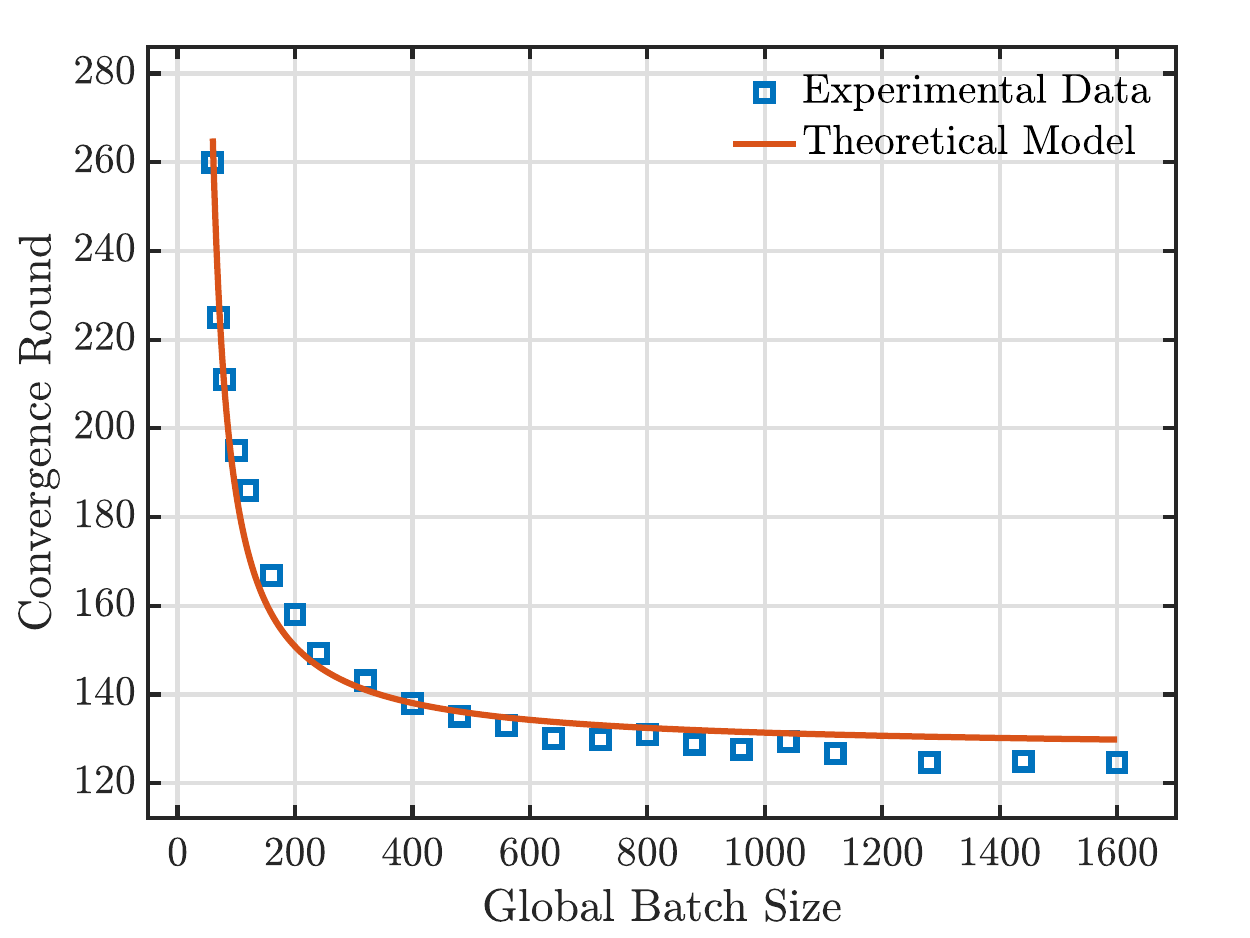}
    \caption{Convergence round vs. global batch size relationship, validated on MNIST with a  \emph{convolutional neural network} (CNN) model (accuracy threshold $\epsilon_A$ = 97\%).}
    \label{fig:round_relationship}
    \vspace{-3mm}
\end{figure}

Fig.~\ref{fig:round_relationship} validates the practicality of the proposed round–batch scaling law, demonstrating a close fit between the theoretical model and empirical results with negligible error.
Additionally, the monotonically decreasing relation between global batch size and convergence round is verified for tradeoff-centric batch-size control in the following subsections.

\subsection{Problem Formulation} 
\label{subsec:fixed_problem_formulation}
In this subsection, we formulate the optimization problem under slow-fading channels where heterogeneous channel states remain constant throughout the learning process. Consequently, the fixed channel coefficients render the communication latency for each device $k$ time-invariant via \eqref{eq:upload_time}, i.e., $T_{k,n}^{\sf{cmm}} = T_{k}^{\sf{cmm}}$ for all $n$.
Accordingly, the per‐round latency $\tau_n$ in \eqref{eq:per_round_latency} is determined by the slowest device, given as
\begin{equation}\label{eq:fixed_latency_tau_n}
  \tau_n=
\max_{k\in\mathcal K}\left(T_k^{\sf{cmm}} + \frac{HWb_{k,n}}{f_k}\right),
\end{equation}
where $\mathcal K= \{1,2,\dots,K\}$.
Building on the estimated batch-round scaling law in \eqref{eq:param_est}, the minimization problem of E2E learning latency can be formulated as
\begin{subequations}
\label{eq:p1}
\begin{align}
 \quad
\min_{N, B, \{b_{k,n}\}} 
& \quad \sum_{n=1}^{N} \tau_n 
\label{eq:p1_obj} \\
\text{s.t.} \quad
& \frac{\hat{\alpha}}{N} + \frac{\hat{\beta}}{B} \le \epsilon, 
\label{eq:convergence_constraint} \\
& \sum_{k=1}^K b_{k,n} = B, \quad \forall n, 
\label{eq:vary_batch_conservation} \\
& N, B, b_{k,n} \in \mathbb{Z}^+, \quad \forall k, n, 
\label{eq:vary_para_domin}
\end{align}
\end{subequations}
where $N$, $B$, and $\{b_{k,n}\}$ represent the number of communication rounds, global batch size, and the batch size of device $k$ in round $n$, respectively.
The constraint \eqref{eq:convergence_constraint} serves as a sufficient condition to guarantee that the parameter pair \((N, B)\) achieves the \(\epsilon\)-level convergence, where \(\hat{\alpha}\) and \(\hat{\beta}\) represent task-specific coefficients estimated in \eqref{eq:param_est}. Moreover,
\eqref{eq:vary_batch_conservation} enforces that the sum of all per-device batch allocations matches the global batch size to ensure convergence, and \eqref{eq:vary_para_domin} constrains $N$, $B$ and $\{b_{k,n}\}$ to be positive integers. 

Note that the combination of discrete decision variables, the nonlinear $\max$-operator in $\tau_n$, and the coupling between $N$ and $B$ renders Problem \eqref{eq:p1} a \emph{mixed‐integer nonlinear program} (MINLP), which is proved to be NP-hard~\cite{garey2002computers}. 
To address this optimization problem, we propose a two-stage optimization approach and derive the optimal batch-size control in the next subsection.

\subsection{Optimal Batch-Size Control}

In this subsection, we solve Problem \eqref{eq:p1} by decoupling it into two independent subproblems.
The decoupling results from two observations under a given global batch size.  One is that the per-round latency $\tau_n$ depends solely on device batch allocations $\{b_{k,n}\}$, as demonstrated in \eqref{eq:fixed_latency_tau_n}.
The other is that the convergence round $N_\epsilon$ is a function of $B$, which can be obtained by rearranging \eqref{eq:convergence_constraint} and taking an integer, given by
\begin{equation}
    N_\epsilon(B) = \left\lceil \frac{\hat\alpha}{\epsilon - \hat\beta/B} \right\rceil,
    \label{eq:N_B}
\end{equation}
where $\lceil x \rceil$ denotes the smallest integer greater than or equal to $x$. This formulation  requires $B > \frac{\hat\beta}{\epsilon}$ to ensure $N_\epsilon(B)$ is a positive integer, a condition maintained throughout subsequent analysis.
Therefore,  we optimize the local batch size $\{b_{k,n}\}$ to minimize the per-round latency $\tau_n$ under a given fixed $B$ in the first stage, followed by the optimization of $B$ to balance the C$^2$ tradeoff. The two steps are elaborated as follows.

\subsubsection{Per-Round Latency Minimization} 
\label{subsec:per_round_optimization}

Consider round $n$ requiring a fixed global batch size $B$, and the minimization of $\tau_n$ can be formulated as
\begin{equation}
\label{eq:p2}
\begin{aligned}
 \quad 
\min_{\{b_{k,n}\}} &\quad \tau_n 
\\
\text{s.t.} &\quad 
\eqref{eq:vary_batch_conservation}, 
\quad b_{k,n} \in \mathbb{Z}^+, \; \forall k, n.
\end{aligned}
\end{equation}
For tractability, we relax the integrality constraint of \(b_{k,n}\in\mathbb Z^+\) to a continuous domain $\tilde b_{k,n} \in [1, B-K+1]$.  The relaxed per‐round latency becomes
    $\tilde{\tau}_n=\max_{k\in\mathcal K}\left (T_k^{\sf{cmm}} + \frac{HW\tilde{b}_{k,n}}{f_k}\right)$, leading to the continuous optimization problem
\begin{subequations}
\label{eq:p3}
\begin{align}
    \min_{\{\tilde{b}_{k,n}\}} \quad
      &\tilde{\tau}_n\\
    \text{s.t.}\quad
    &\sum_{k=1}^K \tilde{b}_{k,n} = B\label{eq:continuous_batch_conservation}, \quad \forall n ,\\
    &1 \le\tilde b_{k,n}\le B-K+1, \quad \forall k, n.
\end{align}
\end{subequations}

Since Problem \eqref{eq:p3} is a min–max optimization over device‐specific latencies, its optimum is attained when each device’s individual latency contributes equally to that maximum, as formalized in the following Lemma~\ref{lemma:latency_equilibrium}.

\begin{Lemma}[Latency Equilibrium Principle] 
\label{lemma:latency_equilibrium}
Let $B^{\sf th}$ denote the threshold of the global batch size, given by
\begin{equation}\label{eq:B_threshold}
B^{\sf th} = \sum_{k=1}^K  
\left\lceil
\frac{f_k}{HW} \left( \tau_{\sf{1b}} - T_k^{\sf{cmm}} \right)
\right\rceil,
\end{equation}
where $\tau_{\sf{1b}}$ denotes the per-round latency with each device processing one batch, given by
\begin{equation}
\tau_{\sf{1b}} = \max_{k \in \mathcal{K}} \left( T_k^{\sf{cmm}} + \frac{HW}{f_k} \right).
\label{eq:taumax_def}
\end{equation}
Conditioned on \(B > B^{\sf th}\), the optimal solutions of Problem \eqref{eq:p3}, which are denoted by $\{\tilde{b}_{k,n}^\star\}$, satisfy the
latency equilibrium, given by 
\begin{equation}
T_i^{\sf{cmm}} + \frac{HW \tilde{b}_{i,n}^\star}{f_i}
= 
T_j^{\sf{cmm}} + \frac{HW \tilde{b}_{j,n}^\star}{f_j} = \tau_{\min}(B),
\; \forall i,j \in \mathcal{K},
\label{eq:latency_equivalence}
\end{equation}
where $\tau_{\min}(B)$ denotes the minimum  per-round latency for global batch size $B$.
\end{Lemma}

\noindent\textit{Proof:} See Appendix~\ref{subsec:latency_balance_proof}.


Rearranging \eqref{eq:latency_equivalence}  gives the optimal local batch size, given by
\begin{equation}
\tilde{b}_{k,n}^\star = \frac{f_k}{HW}\left(\tau_{\min}(B) - T_k^{\sf{cmm}}\right), \forall n.
\label{eq:allocation_principle}
\end{equation}
Note that the fixed communication latency results in a time-invariant local batch size\footnote{Since the local batch size must be an integer, $\{\tilde{b}_{k,n}^\star\}$ is rounded to the nearest integer in practice. Experimental results in Sec.~\ref{sec:experimental results} demonstrate that the performance gap introduced by this rounding is negligible.}. 


Substituting \eqref{eq:allocation_principle} into the batch conservation constraint \eqref{eq:continuous_batch_conservation} yields an explicit expression of the minimum per-round latency $\tau_{\min}(B)$ in the case of all $B > B^{\sf th}$, given by
\begin{equation}
\tau_{\min}(B) = \frac{HWB + \hat{f}_\Sigma}{f_{\Sigma}}, 
\label{eq:system_latency}
\end{equation}
where $f_{\Sigma} =\sum_{k=1}^K f_k$ denotes the aggregate computation capacity across all devices and $\hat{f}_\Sigma = \sum_{k=1}^K f_k T_k^{\sf{cmm}}$ represents the system’s C$^2$ coupling factor.

On the other hand, when the global batch size $B$ is insufficient to support latency balancing across all devices, i.e., $B \leq B^{\sf th}$,
the per-round latency is constrained by the maximum single-batch processing time among all devices due to the integer batch assignment ($b_{k,n} \geq 1$). Thus the minimum achievable per-round latency is given by
\begin{equation}
\tau_{\min}(B) = \tau_{\sf{1b}},
\label{eq:system_latency_straggler}
\end{equation}
where $\tau_{\sf{1b}}$ is defined in \eqref{eq:taumax_def}.



\begin{figure}[t]
    \centering
    \includegraphics[width=0.7\columnwidth]{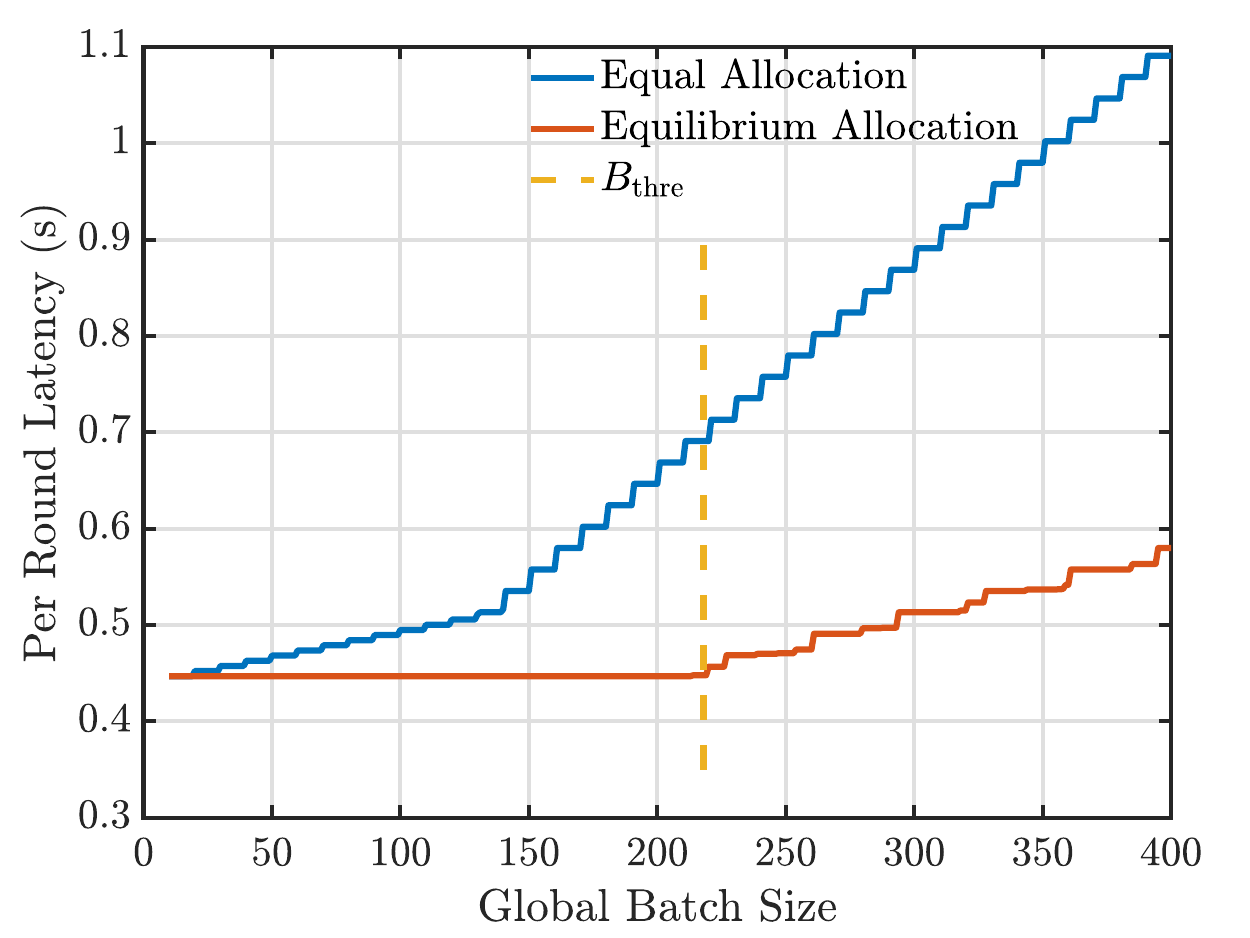}
    \caption{Per-round latency comparison of proposed allocation method with equal allocation method. }
    \label{fig:allocation_method_comparison}
    \vspace{-3mm}
\end{figure}

Fig.~\ref{fig:allocation_method_comparison} compares the per-round latency of the proposed equilibrium batch allocation with the conventional equal allocation approach. Due to integer-valued $\{b_{k,n}\}$, each increment in $B$ triggers discrete reallocation events, causing stepwise latency fluctuations. 
With the proposed scheme, latency stays flat at $\tau_{\sf{1b}}$ until $B$ reaches the threshold $B^{\sf th}$, after which it scales optimally according to \eqref{eq:system_latency}. This scaling leverages the aggregate computational capacity $f_\Sigma$ of all devices. In contrast, equal allocation leads to latency dominated by the device with the highest communication delay initially, and by the weakest computing device as $B$ grows. These results demonstrate the superior latency performance of the proposed strategy across different global batch sizes.

\subsubsection{C$^2$ Tradeoff Optimization}
\label{subsec:global_coordination}
We optimize the global batch size \(B\) to balance the C$^2$ tradeoff while satisfying the convergence requirement. 
By introducing the obtained convergence round and per‐round latency into \eqref{eq:total_latency}, the E2E latency under time-invariant communication latency, denoted by $\psi(B)$, is computed as
\begin{equation}
\begin{split}
        \psi(B) &= N_{
\epsilon}(B) \cdot \tau_{\min}(B)\\
& =\begin{cases}
  \left\lceil \frac{\hat\alpha}{\epsilon - \hat\beta/B} \right\rceil\cdot\displaystyle \frac{HWB + \hat{f}_\Sigma}{f_{\Sigma}}, & B > B^{\sf th},\\
  \left\lceil \frac{\hat\alpha}{\epsilon - \hat\beta/B} \right\rceil \cdot\displaystyle \tau_{\sf{1b}}, & B \leq B^{\sf th},
\end{cases}
\end{split}
\end{equation}
where $N_{\epsilon}(B)$ defined in \eqref{eq:N_B}  guarantees $\epsilon$-level convergence, and \(\tau_{\min}(B)\) is the piecewise function obtained in \eqref{eq:system_latency} and \eqref{eq:system_latency_straggler} via the optimal local batch-size allocation in \eqref{eq:allocation_principle}.

\begin{figure}[t]
    \centering
    \subfigure[Bandwidth $B_W=1$ MHz\label{fig:E2E_surrogate_comparison1}]{
\includegraphics[width=0.48\columnwidth]{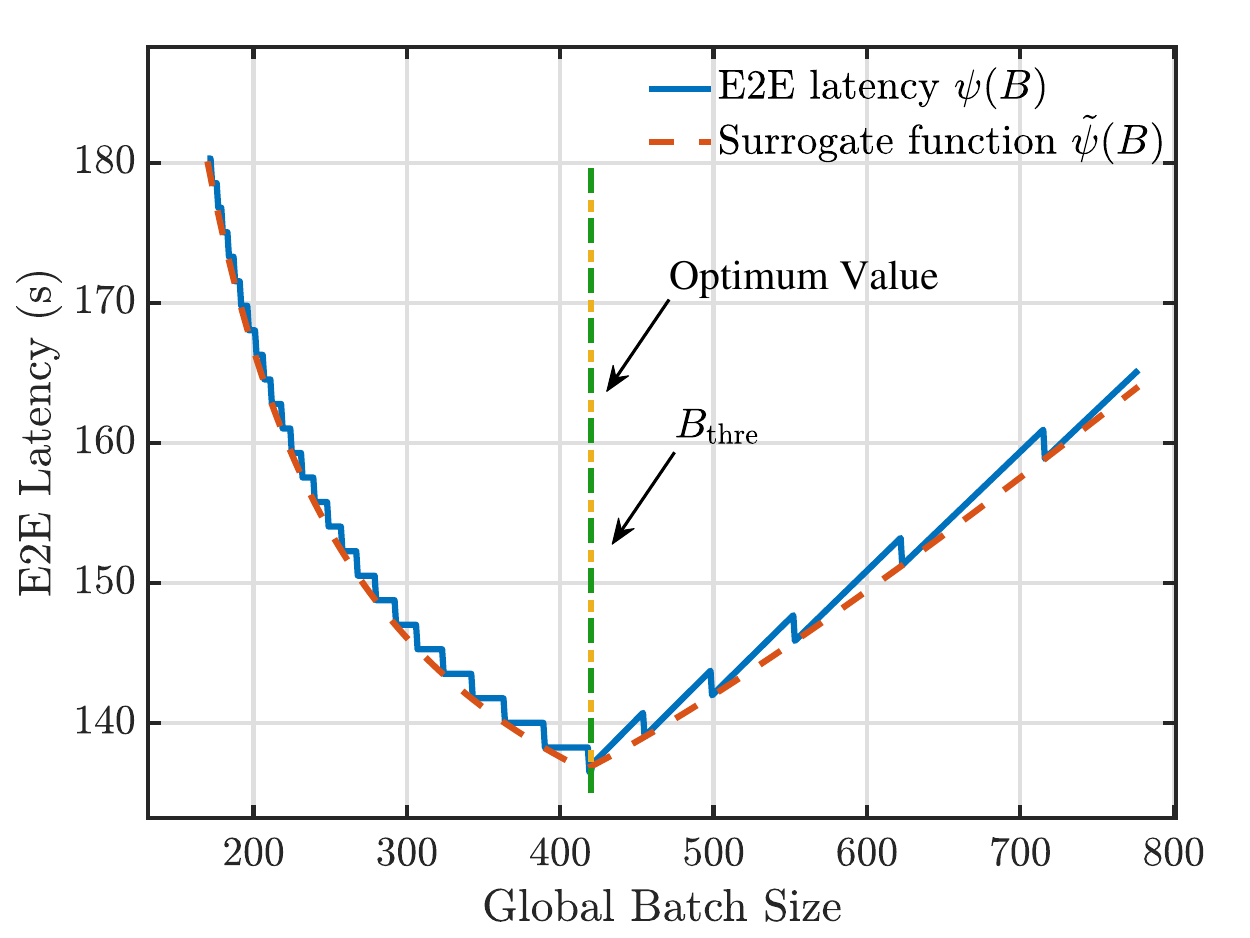}}
    \subfigure[Bandwidth $B_W=2$ MHz\label{fig:E2E_surrogate_comparison2}]{
        \includegraphics[width=0.48\columnwidth]{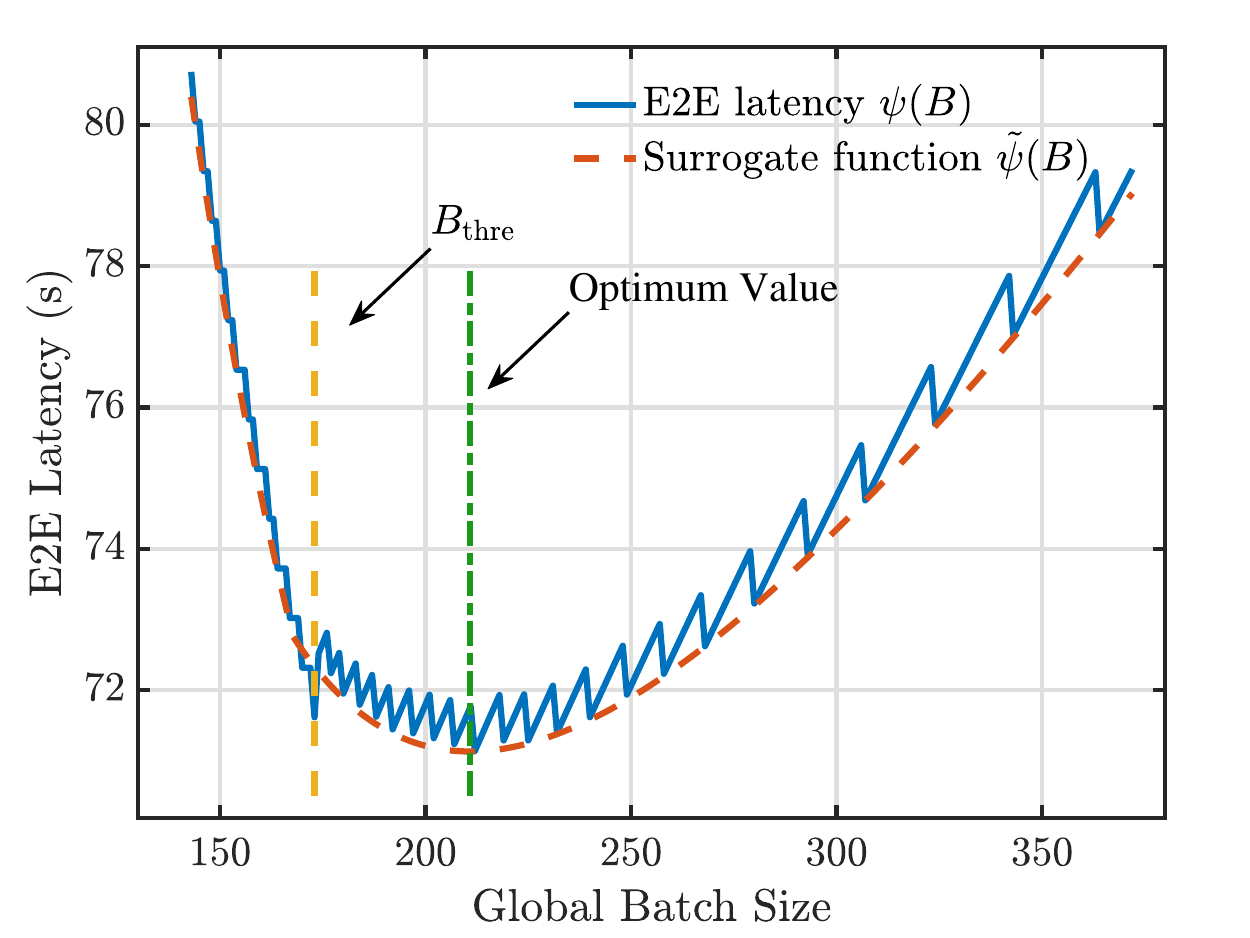}}
    \caption{Two C$^2$ tradeoff regimes through comparison of E2E latency $\psi(B)$ and its continuous relaxation $\tilde{\psi}(B)$ under different bandwidth configurations (CNN model, MNIST dataset).
    }
    \label{fig:E2E_surrogate_comparison}
    \vspace{-3mm}
\end{figure}

In Fig.~\ref{fig:E2E_surrogate_comparison},
the E2E latency 
$\psi(B)$ characterizes the C$^2$ tradeoff governed by global batch size $B$. Specifically, increasing $B$ initially reduces $\psi(B)$ until a minimum point, which comes from the improved gradient quality.
Then  $\psi(B)$ is observed to increase with rise of $B$ due to the lengthened local computation latency.
The jaggedness of $\psi(B)$ arises from the ceiling operation on $N_{\epsilon}(B)$.

For tractable optimization, we consider a continuous relaxation of $\psi(B)$ by allowing $N_{\epsilon}(B)$ and $B$ to take non-integer values~\cite{wang2024ultra}.
 This yields the continuous surrogate function of $\psi(B)$, denoted as $\tilde{\psi}(B)$, given by
\begin{equation}\label{eq:continuous_overall_E2E_latency}
\tilde{\psi}(B) =
\begin{cases}
  \dfrac{\hat{\alpha} B \left(H W B + \hat{f}_{\Sigma} \right)}{f_{\Sigma}  \left( \epsilon B - \hat{\beta} \right)} , & B > \tilde{B}^{\sf th}, \\
   \dfrac{\hat{\alpha} B \tau_{\sf{1b}}}{\epsilon B - \hat{\beta}} , & B \leq \tilde{B}^{\sf th},
\end{cases}
\end{equation}
where
$\tilde{B}^{\sf th}=\sum_{k=1}^K 
\left(
\frac{f_k}{HW} \left( \tau_{\sf{1b}} - T_k^{\sf{cmm}} \right)
\right)$ is the adjusted threshold of global batch size.
Fig.~\ref{fig:E2E_surrogate_comparison} demonstrates that this relaxation preserves monotonicity and closely approximates the exact expression of $\psi(B)$  while capturing the optimal global batch size. 
With such a surrogate, the optimization Problem \eqref{eq:p1} is reformulated as
\begin{equation}
\label{prob:p22}
\begin{split}
    \min_{B} & \quad \tilde{\psi}(B)\\
    \text{s.t.} & \quad B\in \left(\frac{\hat\beta}{\epsilon},B_{\max}\right],
\end{split} 
\end{equation}
where $B>\frac{\hat\beta}{\epsilon}$ ensures a positive convergence round and $B_{\max}$ denotes the maximum feasible global batch size per round.
The value of \(B_{\max}\) is selected sufficiently large so as not to constrain the true optimum.
The objective of Problem \eqref{prob:p22} is proven to be a unimodal function of  $B\in \left(\frac{\hat\beta}{\epsilon}, B_{\max}\right]$, with a unique minimum point. The optimal solution is formalized in Theorem \ref{thm:optimal_coordination}.

\begin{Theorem}[Optimal Batch-Size Control]
\label{thm:optimal_coordination}
Considering the heterogeneous FL system with slow-fading channels, the optimal global batch size $B^*$ that solves Problem \eqref{eq:p1} is given by
\begin{equation}
\label{eq:B_optimal}
    B^* = \max\left\{ B^{\sf th},\; \left\lfloor B_{\epsilon}\right\rceil_{\tilde{\psi}(\cdot)} \right\},
\end{equation}
where $B_{\epsilon} = \frac{\hat{\beta}}{\epsilon} 
    \left( 1 + \sqrt{1 + \frac{\hat{f}_{\Sigma} \epsilon}{HW \hat{\beta}}} \right)$ and $\lfloor x \rceil_{\tilde{\psi}(\cdot)}$ is equal to $\lfloor x \rfloor$ if $\tilde{\psi}(\lfloor x \rfloor) \leq \tilde{\psi}(\lceil x \rceil)$, and is otherwise equal to $\lceil x \rceil$.

The corresponding device-level allocations are given as
\begin{equation}
\label{eq:device_allocation}
    b_{k,n}^* = \operatorname{round}\left( \frac{f_k}{HW}  \left( \frac{HW B^* + \hat{f}_\Sigma}{f_{\Sigma}} - T_k^{\sf{cmm}} \right) \right), \; \forall k,n,
\end{equation}
where the $\operatorname{round}(\cdot)$ operator denotes rounding to the nearest integer.
\end{Theorem}


\noindent \textit{Proof:} See Appendix \ref{subsec:optimal_batch_proof}.



According to Theorem \ref{thm:optimal_coordination}, \eqref{eq:B_optimal} reveals that the optimal global batch size $B^*$ is attained at either $B^{\sf th}$ or $\left\lfloor B_{\epsilon}\right\rceil_{\tilde{\psi}(\cdot)}$, as depicted in Fig.~\ref{fig:E2E_surrogate_comparison1} and Fig.~\ref{fig:E2E_surrogate_comparison2}, respectively. The max operator in \eqref{eq:B_optimal} ensures $B^* \ge B^{\sf th}$, guaranteeing sufficient samples per device to achieve synchronized completion and eliminate idle time. 
Furthermore, $B^*$ increases with the C$^2$ competing factor $\hat{f}_{\Sigma}$ that indicates the superiority of computation over communication capabilities.
Specifically, a larger $\hat{f}_{\Sigma}$ indicates that systems with either higher computational capabilities (larger $f_k$) or longer communication delays (larger $T_k^{\sf{cmm}}$) necessitate a larger global batch size to minimize E2E latency, and vice versa.
Complementing the global optimization, the device-level allocations $\{b_{k,n}^*\}$ adapt to heterogeneous capabilities via the linear relationship in \eqref{eq:device_allocation}. Devices with higher $f_k$ or lower $T_k^{\sf{cmm}}$ receive proportionally larger batches, avoiding resource waste when waiting for stragglers.

Since both $B^*$ and $\{b_{k,n}^*\}$ are closed‑form expressions requiring only simple arithmetic across the $K$ devices, the total computational complexity grows linearly as $\mathcal{O}(K)$. This constitutes an exponential improvement over general‑purpose MINLP solvers (e.g., CPLEX~\cite{cplex}, Gurobi~\cite{gurobi}), whose runtimes typically scale exponentially with the number of integer variables~\cite{garey2002computers}. Thus, our analytical solution incurs negligible computational overhead and enables real‑time coordination in large‑scale FL deployments.

\section{Adaptive Batch-Size Control with fast fading}
\label{sec:Adaptive_Batch_Size_Control}

The preceding section focuses on the FL system with slow-fading channels. In this section, we extend the derived optimal control framework to design channel-adaptive batch sizes that adapt to fast-fading channels and minimize E2E learning latency.

\subsection{Problem Formulation}
\label{subsec:adaptive_problem_formulation}

This subsection extends the analysis to fast-fading channels that vary across rounds. Consequently, time-varying channel coefficients lead to round-dependent communication latency, denoted by \(T^{\sf{cmm}}_{k,n}\) for device \(k\) in round \(n\). Under this setting, the edge server dynamically determines each device’s batch size $b_{k,n}$ in every round.
The resulting E2E latency minimization problem in this case is formulated by Problem \eqref{eq:p1}, where per-round latency $\tau_n$ is redefined  by  
\begin{equation}
    \tau_n=\max_{k \in \mathcal{K}} \left( T^{\sf cmm}_{k,n} + \frac{HW b_{k,n}}{f_k} \right).
\end{equation}

Unlike the minimum per-round latency in \eqref{eq:system_latency} and \eqref{eq:system_latency_straggler} that remains fixed over rounds, the effects of $\tau_n$ on the E2E latency are unpredictable due to independent realization of channel over rounds. This makes the derivation of an optimal batch-size control policy particularly challenging.
To address this limitation, we propose a suboptimal approach that adapts batch-size control based on the statistical characteristics of dynamic channels, as detailed in the following subsection.



\subsection{Adaptive Batch-Size Control}
\label{ssec:adaptive_batch}

\begin{figure}[t]
    \centering
    \includegraphics[width=0.7\columnwidth]{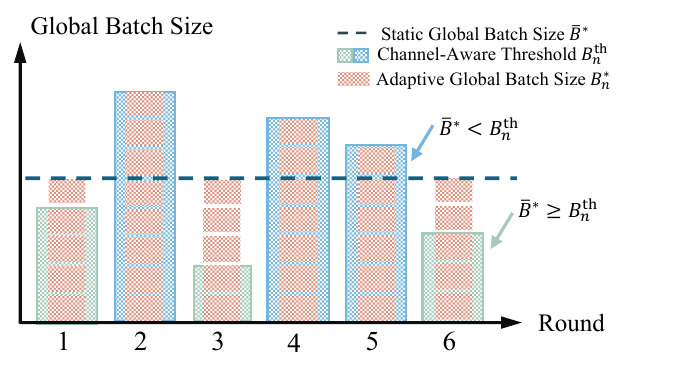}
    \caption{Adaptive batch-size control mechanisms.}
    \label{fig:two_case_B}
    \vspace{-3mm}
\end{figure}
In this subsection, we first compute a static global batch size based on long-term channel statistics to balance the C$^2$ tradeoff. Then, we adapt the batch-size control to the instantaneous channel realizations in each round.

\subsubsection{Static Global Batch Size}  
\label{sssec:long_term_opt}

This step characterizes the effects of long-term channel states on the C² tradeoff. We consider a static global batch size, denoted by \(\bar{B}^*\), which is computed by substituting the expected communication latency \(\bar{T}^{\mathrm{cmm}}_{k} = \mathbb{E}[T^{\mathrm{cmm}}_{k,n}]\) into \eqref{eq:B_optimal}. The expectation is derived from the long-term statistical distribution of channel gains, obtained either through offline measurements over historical rounds or analytical derivation from the underlying channel model.
Mathematically, 
\begin{equation}
\label{eq:B_optimal_dyn}
    \bar{B}^* = \max\left\{\bar{B}^{\sf th},\left\lfloor \frac{\hat\beta}{\epsilon} \left( 1 + \sqrt{1 + \frac{\hat{f}_\Sigma^{\sf{E}} \epsilon}{HW \hat\beta}} \right)\right\rceil_{\tilde{\psi}(\cdot)} \right\},
\end{equation}
where $\hat{f}_\Sigma^{\mathrm{E}} = \sum_{k=1}^K f_k\,\bar{T}^{\mathrm{cmm}}_{k}$ represents the expected communication latency weighted by computation speeds, and $\bar{B}^{\sf th}$ denotes the global batch size threshold with expected communication latencies, given by
\begin{equation}
    \bar{B}^{\sf th}
= \sum_{k=1}^K 
\left\lceil \frac{f_k}{HW}\left(\max_{j \in \mathcal{K}}\left(\bar{T}^{\sf{cmm}}_{j} + \frac{HW}{f_j}\right)
- \bar{T}^{\sf{cmm}}_{k}\right)\right\rceil.\label{eq:Bthre_exp}
\end{equation}
With the C$^2$ tradeoff balanced via the static global batch size $\bar{B}^*$, the channel-aware batch-size adaptation per round is detailed below.

\subsubsection{Channel-Aware Batch-Size Adaptation}  
\label{sec:online_adapt}

To adapt the batch size to current channel realization, we consider the comparison between two key batch size values: 1) the static global batch size $\bar{B}^*$  in \eqref{eq:B_optimal_dyn}, and 2) the channel-aware global batch size threshold $B_n^{\sf th}$ required in round $n$ to realize the latency equilibrium under observed latencies $T^{\sf cmm}_{k,n}$.
It is computed by replacing $T^{\sf cmm}_k$ with $T^{\sf cmm}_{k,n}$ in \eqref{eq:B_threshold}, given by
\begin{equation}
\label{eq:Bmin_dyn}
    B_n^{\sf th} \triangleq\sum_{k=1}^K \left\lceil \frac{f_k}{HW}  \left(  \max_{j \in \mathcal{K}  } \left( T^{\sf cmm}_{j,n} + \frac{HW}{f_j} \right) - T^{\sf cmm}_{k,n} \right)\right\rceil.
\end{equation}

As depicted in Fig.~\ref{fig:two_case_B}, the batch-size adaptation of round $n$ is determined by two cases. If $\bar{B}^* \geq B_n^{\sf th}$ holds, it means a global batch size $\bar{B}^*$ can achieve a statistically balanced tradeoff between per-round latency and convergence round, while simultaneously maintaining per-round latency equilibrium.
Thus the adaptive global batch size, denoted by $B_n^*$, is set to $B_n^* = \bar{B}^*$.
Conversely, if  \(\bar{B}^* < B_n^{\sf th} \) holds, we raise the global batch size to \(B_n^{\sf th}\). This adjustment keeps the round‐specific latency $\tau_n$ unchanged as it remains equal to the maximum straggler delay. At the same time, the increased batch size accelerates convergence by reducing the required rounds (via diminishing $\hat\beta K/B$ term in \eqref{eq:N_B}), thereby further lowering E2E latency.
Accordingly, the adaptive global batch size $B_n^*$ is formulated as
\begin{equation}\label{eq:adaptive_B}
    B_n^* = \max\left(\bar{B}_n^*,B_n^{\sf th}\right).
\end{equation}
Finally, substituting $B_n^*$ and the round-specific $T^{\sf cmm}_{k,n}$ into the allocation rule \eqref{eq:device_allocation} yields the per‐device allocation
\begin{equation}
\label{eq:b_alloc}
    b_{k,n}^* = \operatorname{round}\left( \frac{f_k}{HW} \left( \frac{HW B_n^* + \hat{f}_{\Sigma,n}}{f_{\Sigma}} - T^{\sf cmm}_{k,n} \right) \right),
\end{equation}
where $\hat{f}_{\Sigma,n} = \sum_{k=1}^K f_k T^{\sf cmm}_{k,n}$ and $f_{\Sigma} = \sum_{k=1}^K f_k$.
The complete optimization procedure is formalized in Algorithm~\ref{alg:dynamic_batch},  which outlines the sequential steps involved in achieving adaptive batch-size control.

\begin{algorithm}[t]
\caption{Adaptive Batch-Size Control Protocol}
\label{alg:dynamic_batch}
\begin{algorithmic}[1]
\renewcommand{\algorithmicrequire}{\textbf{Input:}}
\REQUIRE 
Communication parameters: $B_W$, $\{P_k\}$, $N_0$, $q$, $\{h_{k,n}\}$;  
Computation parameters: $W$, $\{f_k\}$;

\STATE Compute $\bar{T}^{\sf{cmm}}_k$ for each device  
\STATE Compute reference batch size $\bar{B}^*$ via \eqref{eq:B_optimal_dyn}

\FOR{each communication round $n=1,2,\dots$}
    \STATE Measure instantaneous latencies $\{T^{\sf{cmm}}_{k,n}\}$  
    \STATE Compute channel-aware threshold $B_n^{\sf th}$ via \eqref{eq:Bmin_dyn}
    \STATE Compute adaptive batch size $B_n^*$ via \eqref{eq:adaptive_B}
    \STATE Compute per-device allocation $b_{k,n}^*$ via \eqref{eq:b_alloc}
\ENDFOR

\RETURN Per‐round allocations $\{b_{k,n}^*\}$ for each round $n$
\end{algorithmic}

\end{algorithm}

\section{Experimental Results}
\label{sec:experimental results}
\subsection{Experimental Settings}  
The experimental configuration adheres to the following specifications unless explicitly stated otherwise.  

\begin{itemize}  
    \item \textbf{Communication Settings:}  
    We consider the FL system comprising one server and $K=10$ heterogeneous edge devices. The total available frequency spectrum is divided into $K$ orthogonal subchannels, each allocated to one device with a bandwidth of $B_W = 10$ MHz. The model parameters are quantized into $Q = 32$ bits and the noise power spectral density is $N_0 = 10^{-10}$ W/Hz.
    The communication heterogeneity across devices is simulated by two aspects: 1) the transmit power $P_k$ of device $k$ is randomly drawn from a uniform distribution over the interval $[0.01, 0.1]$ W; and 2) the channel coefficient $h_{k,n}$ of device $k$ at round $n$ follows an i.i.d. Rayleigh fading model $\mathcal{CN}(0, \sigma_k^2)$, where the scale parameter $\sigma_k^2$ is uniformly sampled from the interval $[0.2, 0.5]$.

    \item \textbf{Learning Settings:}  
    The image classification task is evaluated on two widely used benchmarks: MNIST~\cite{lecun1998gradient}, consisting of 10 grayscale digit classes with 60,000 training and 10,000 validation samples, and CIFAR-10~\cite{krizhevsky2009learning}, comprising 10 color object categories with 50,000 training and 10,000 validation images.
    For MNIST, we employ a CNN with two successive $5\times 5$ convolutional layers (10 and 20 feature maps, respectively), each followed by $2\times 2$ max‑pooling and ReLU activation. A dropout layer ($p=0.5$) follows the second convolution, then a 50-unit fully connected layer (ReLU activation, dropout $p=0.5$) and the output layer, totaling 21840 trainable parameters~\cite{wang2024spectrum}.
    For CIFAR‑10, we adopt ResNet‑18~\cite{he2016deep} with batch‑normalization layers, comprising approximately 11 million parameters. The training data are evenly partitioned into 10 i.i.d. subsets across devices. 
    To emulate heterogeneous computing capabilities across devices, we sample each device’s processing capacity randomly from  $[0.001,0.03] \times \hat{f}$, where $\hat{f}= 1$ TFLOPs/s is the computation speed of an NVIDIA Jetson TX2 Series~\cite{wang2025revisiting}. 
    We conduct $H=5$ local updates per round for a CNN on MNIST and $H=3$ for a ResNet on CIFAR-10, both using a fixed learning rate of $\eta = 0.1$. The target validation accuracy thresholds are set to 95\% of the maximum achievable accuracy, i.e., $\epsilon_A = 95\%$ for MNIST and $\epsilon_A = 58\%$ for CIFAR-10, respectively. Conditioned on these thresholds, we set the convergence tolerance $\epsilon = 0.5$, and estimate the parameters as $\hat\alpha = 34.5$, $\hat\beta = 23.2$ for MNIST, and $\hat\alpha = 30.0$, $\hat\beta = 123.3$ for CIFAR-10.

\end{itemize}  

\begin{figure}[t]
    \centering
    \subfigure[MNIST]{       \includegraphics[width=0.48\columnwidth]{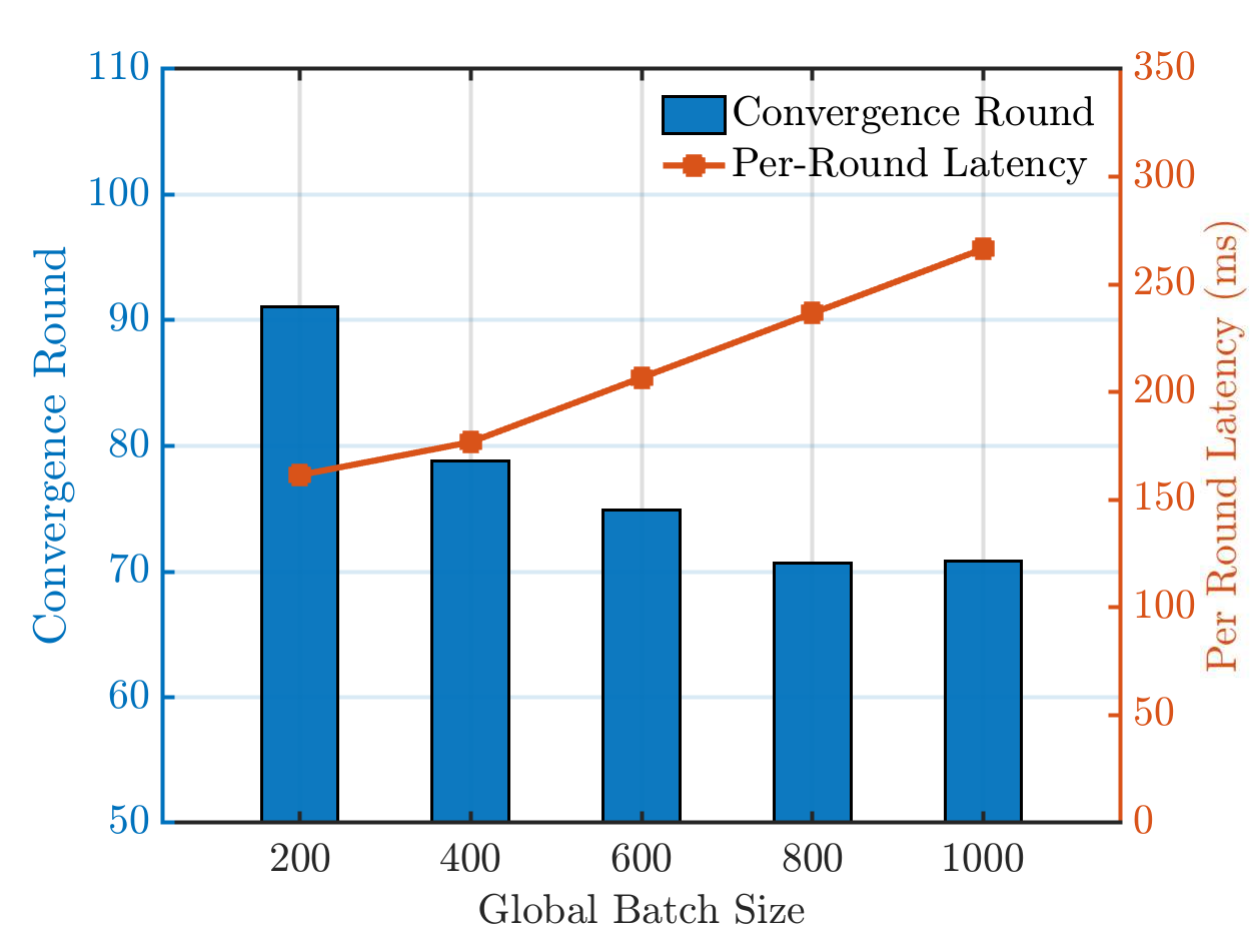}}
    \subfigure[CIFAR-10]{
        \includegraphics[width=0.48\columnwidth]{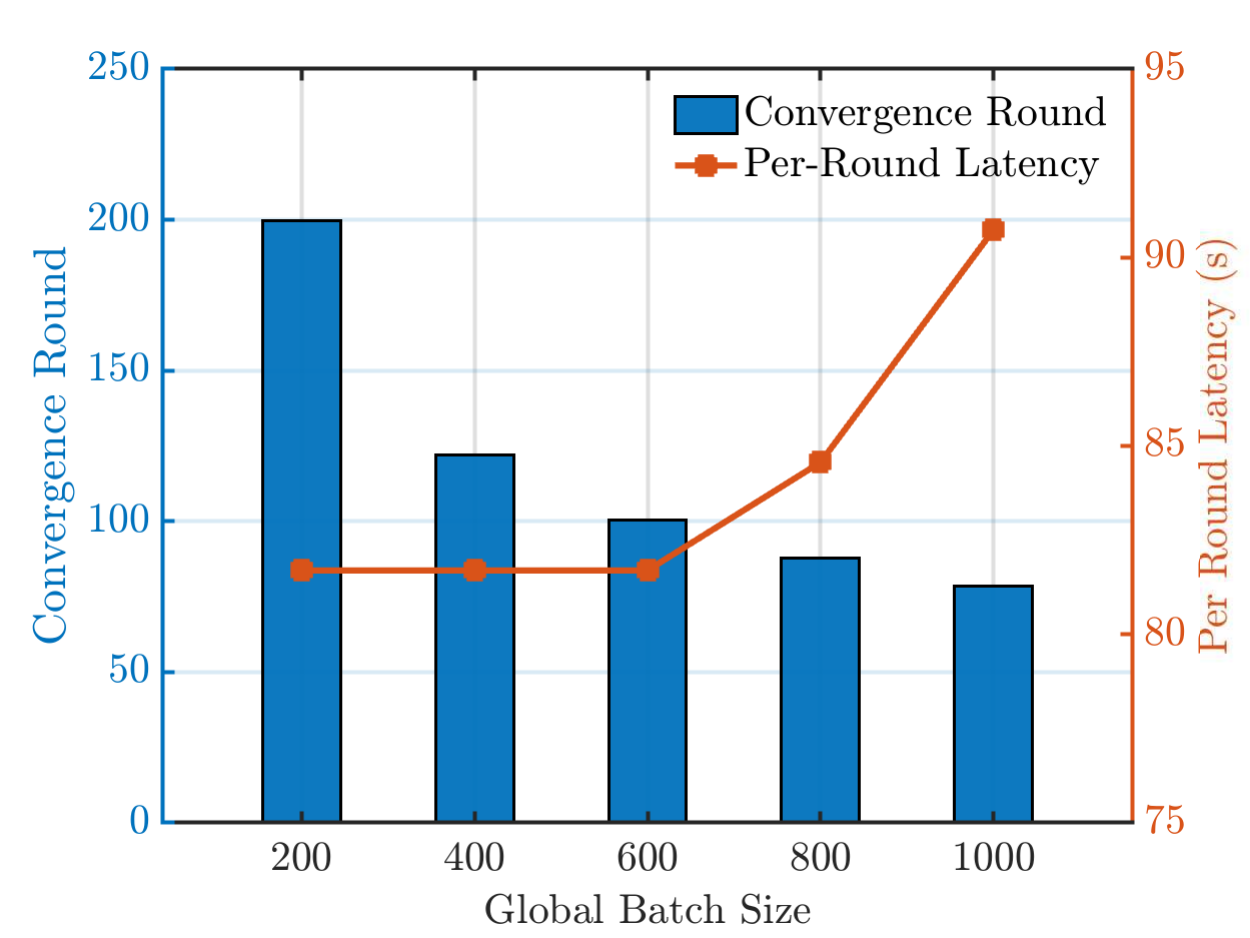}}

    \caption{The convergence round and per-round latency vs. global batch size on both training datasets.}
    \label{fig:convergence_rounds_per_round_latency}
    \vspace{-3mm}
\end{figure}

\begin{figure}[t]
    \centering
    \subfigure[MNIST]{
        \includegraphics[width=0.48\columnwidth]{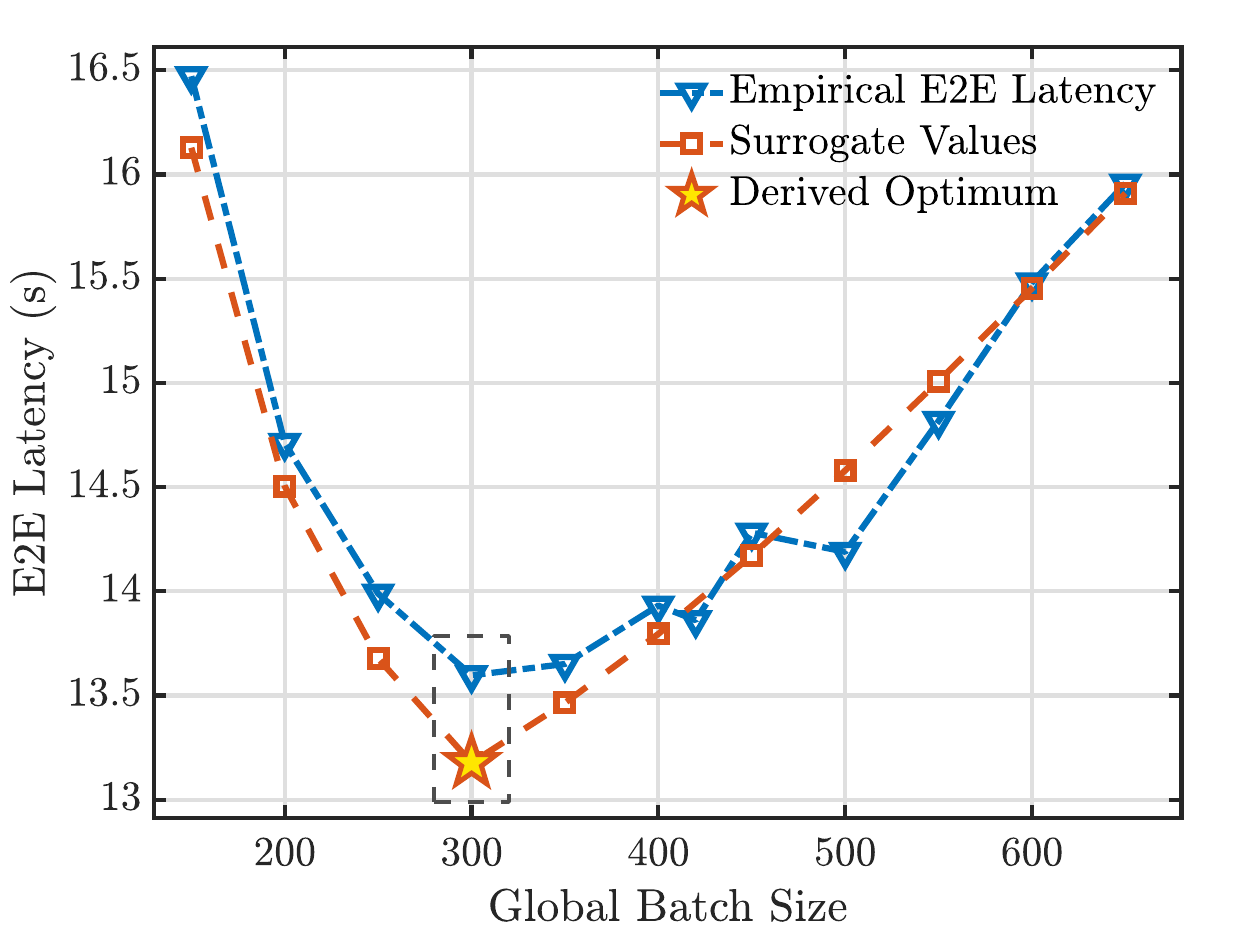}}
    \subfigure[CIFAR-10]{
        \includegraphics[width=0.48\columnwidth]{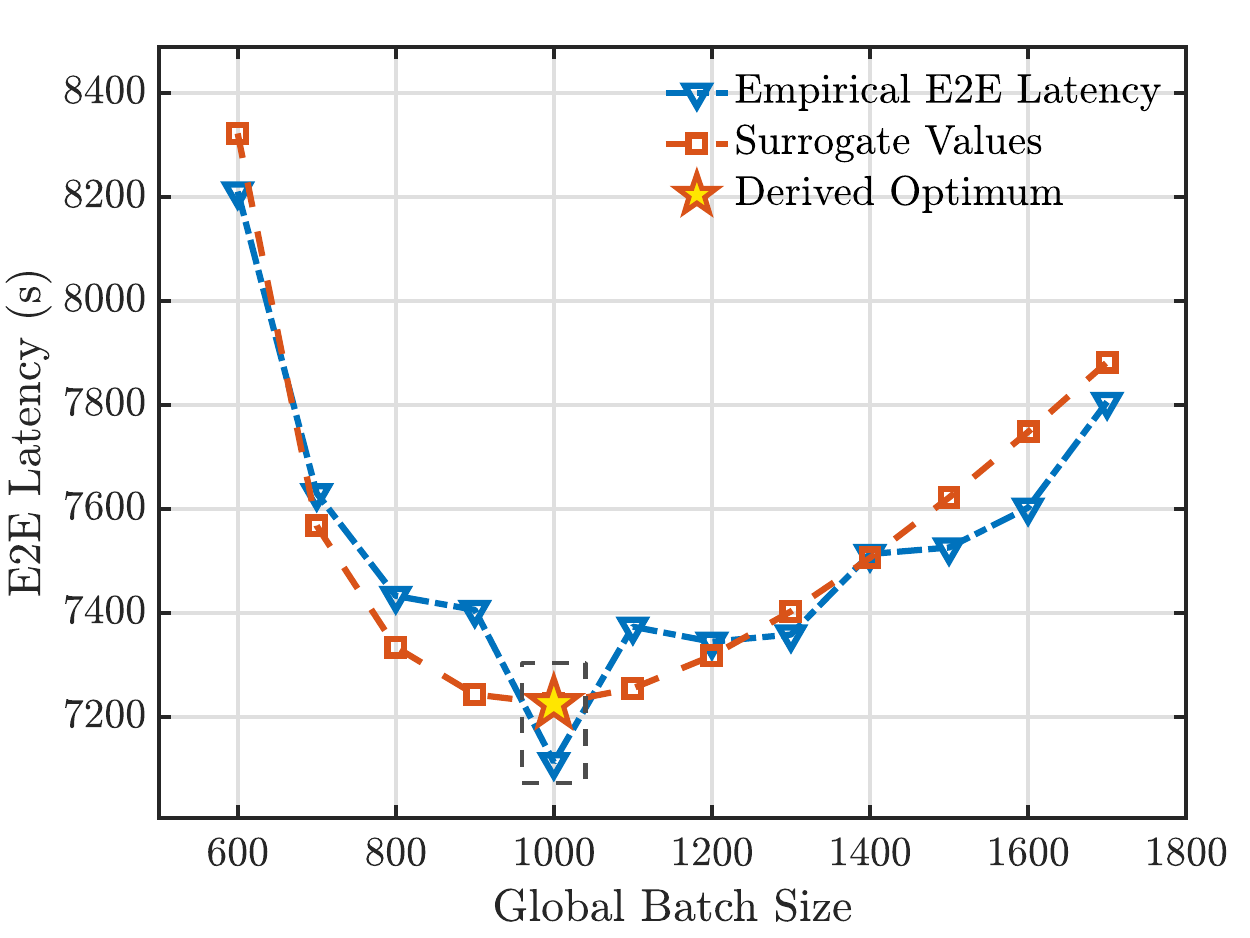}}

    \caption{The illustration of the C$^2$ tradeoff: E2E latency vs. global batch size by empirical measurements, theoretical predictions from the surrogate model in \eqref{eq:continuous_overall_E2E_latency}, and the analytically derived optimum point via \eqref{eq:B_optimal}.}
    \label{fig:practical_vs_theoretical_latency}
    \vspace{-3mm}
\end{figure}

\begin{figure}[t]
    \centering
    \subfigure[MNIST]{
        \includegraphics[width=0.46\columnwidth]{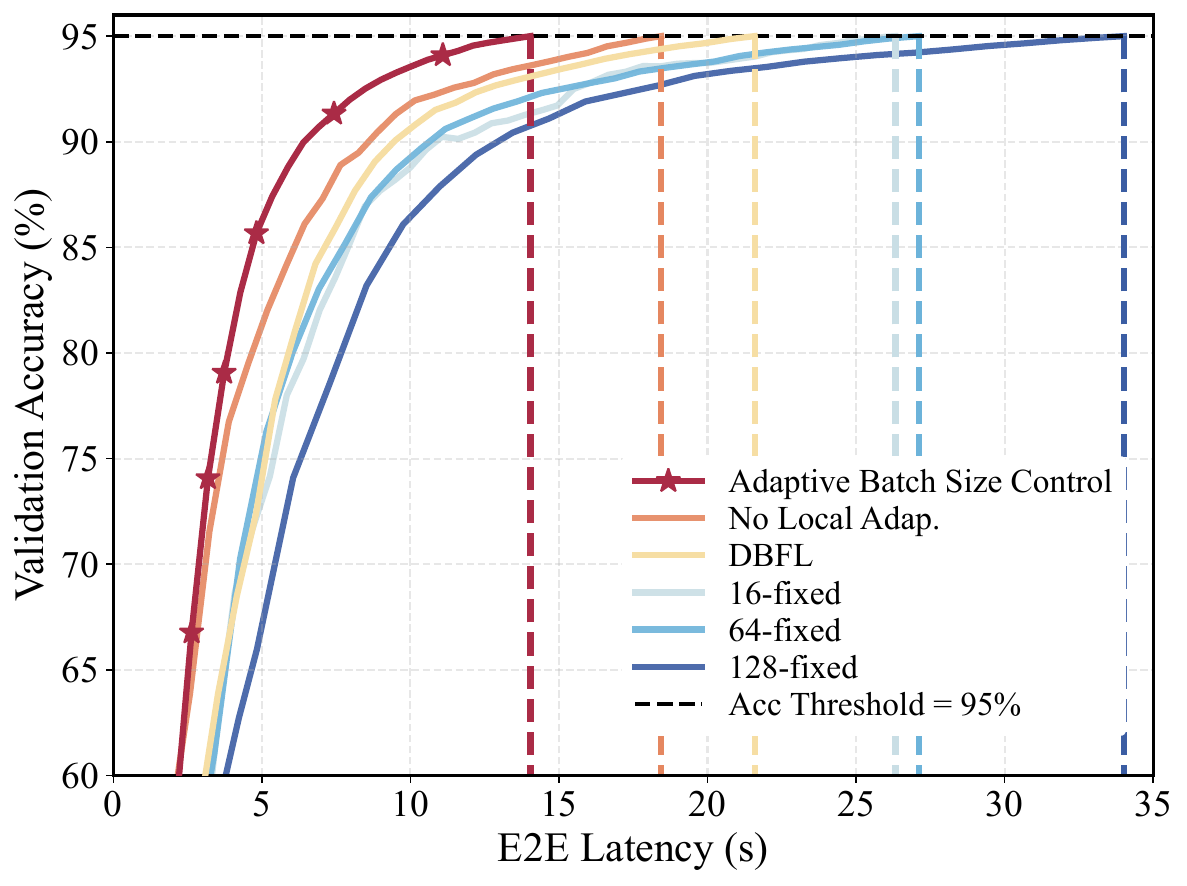}
        \label{fig:MNIST E2E latency vs test accuracy}
    }
    \subfigure[CIFAR-10]{
        \includegraphics[width=0.46\columnwidth]{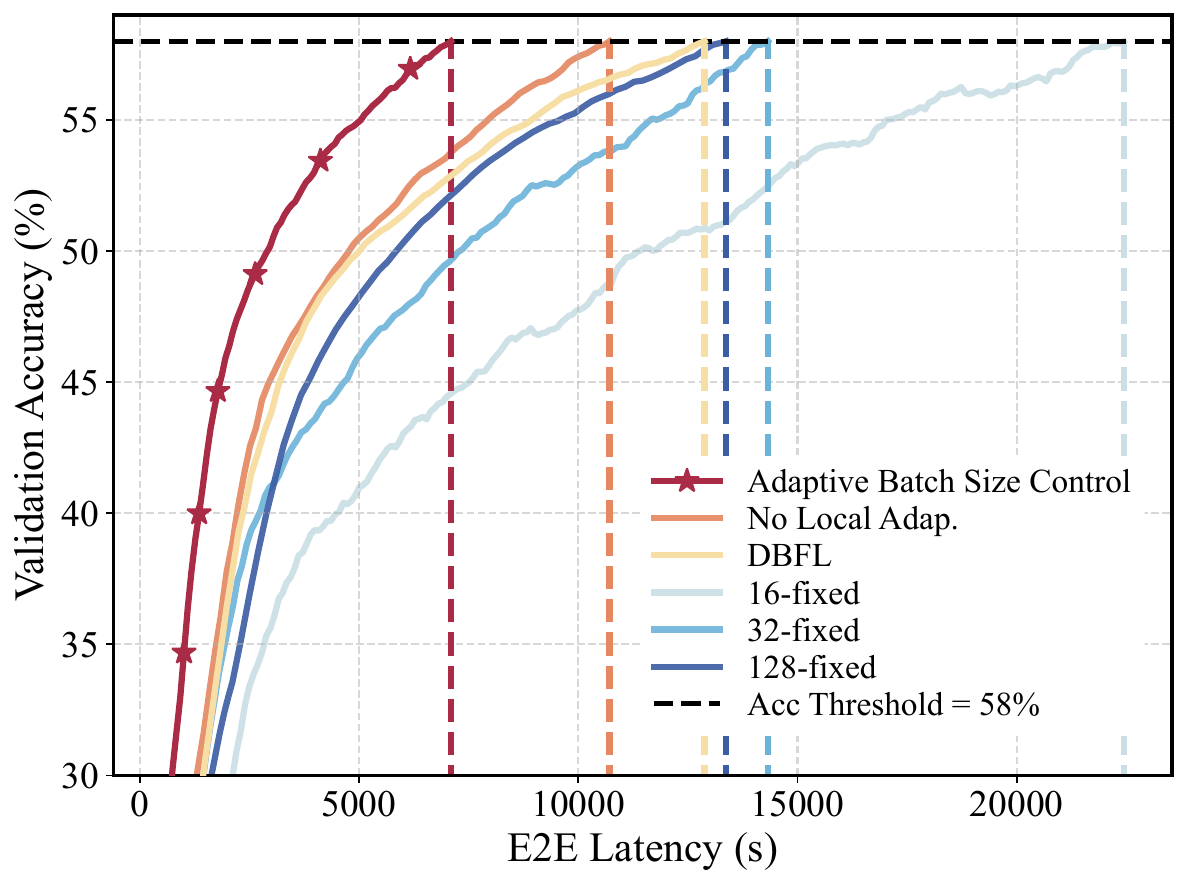}
        \label{fig:CIFAR-10 E2E latency vs test accuracy}
    }
    \caption{Learning performance comparison of different batch control schemes under dynamic communication latency on both datasets. Each scheme terminates upon reaching predefined validation accuracy thresholds.}
    \label{fig:E2E latency vs test accuracy}
    \vspace{-3mm}
\end{figure}
\begin{figure}[t]
    \centering
    \subfigure[MNIST]{
        \includegraphics[width=0.48\columnwidth]{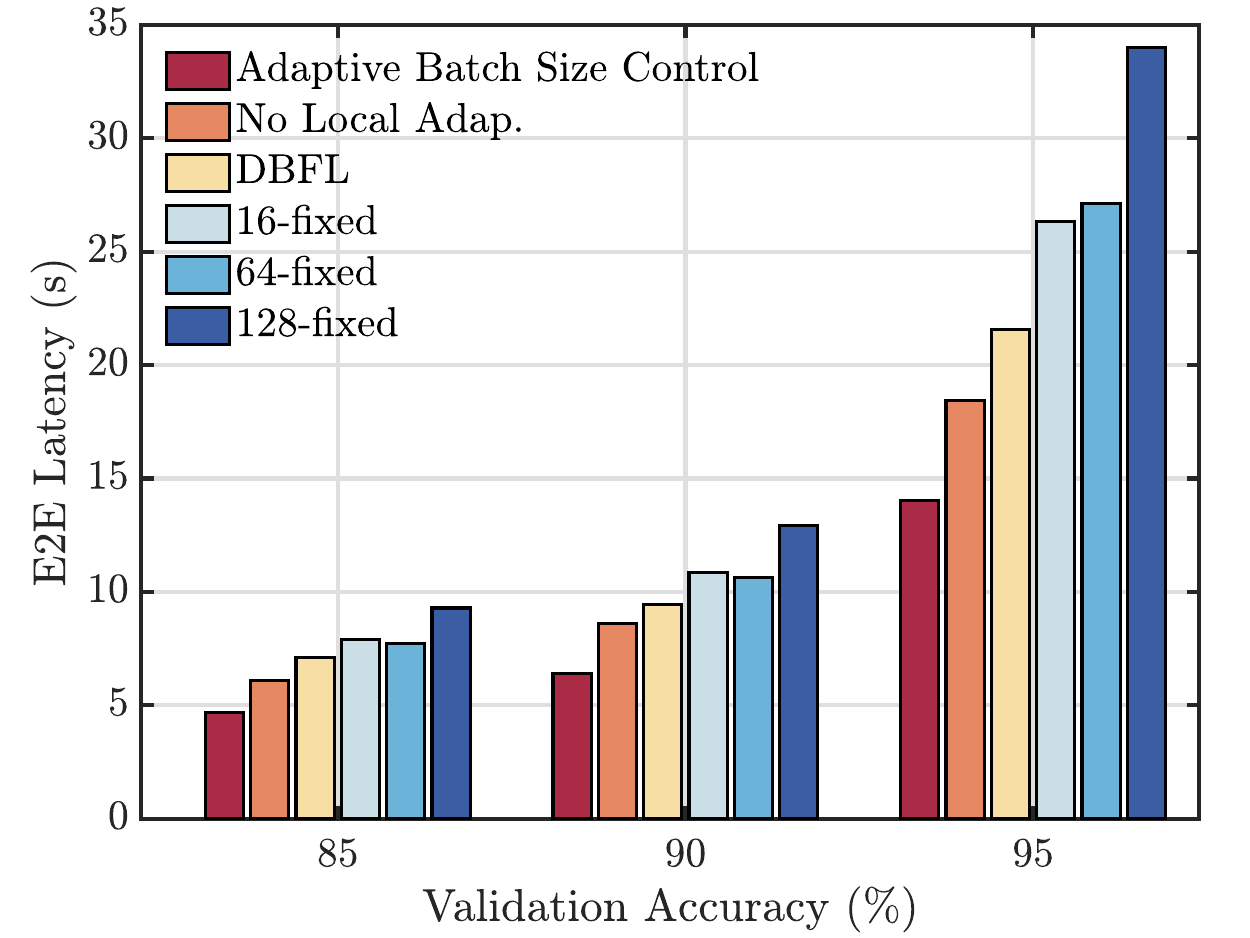}}
    \subfigure[CIFAR-10]{
        \includegraphics[width=0.48\columnwidth]{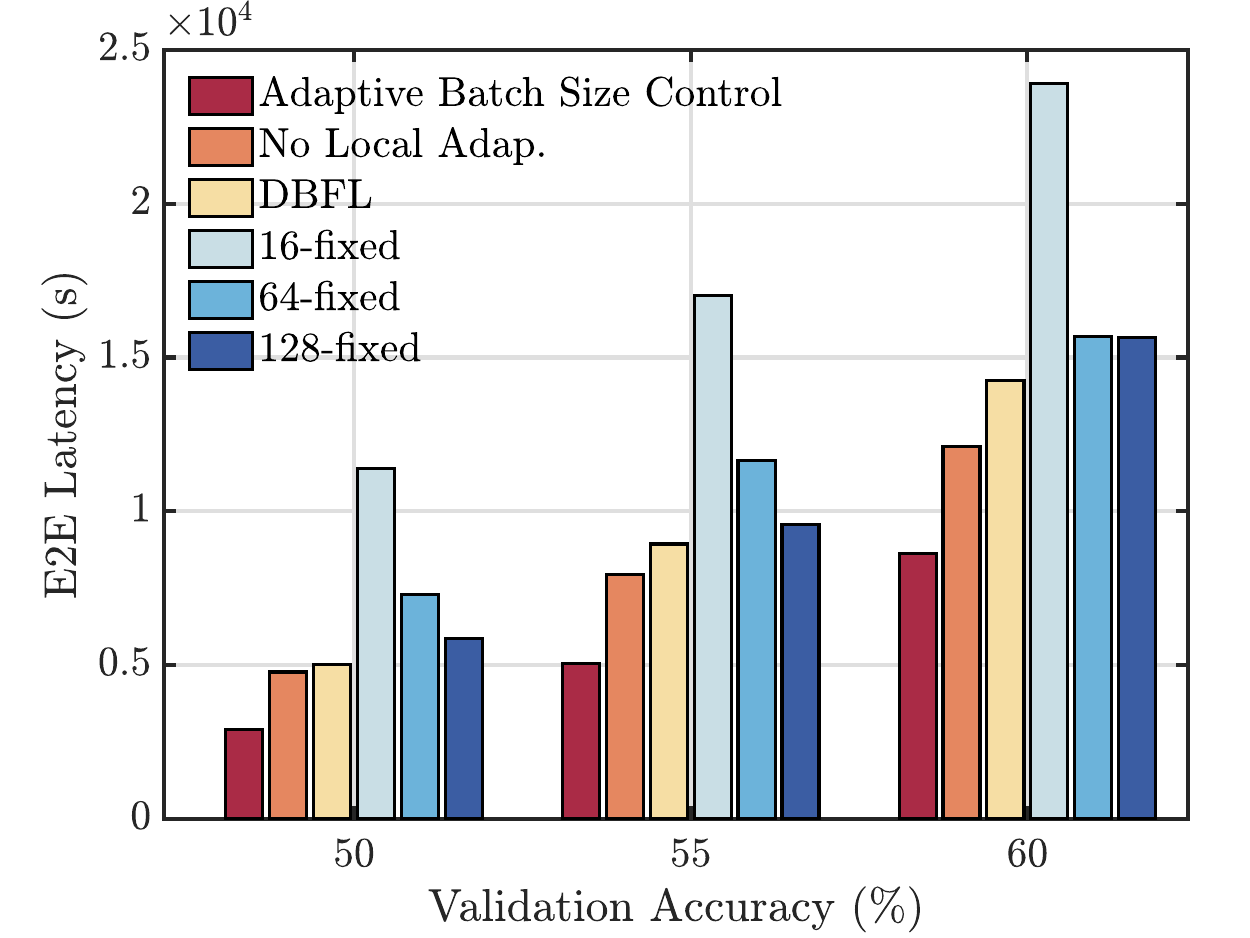}}
    \caption{E2E latency comparison of different batch control schemes for three validation accuracy threshold under dynamic communication latency on both datasets.}
    \label{fig:Accuracy of Different Batch Scheme vs. E2E latency}
    \vspace{-3mm}
\end{figure}

In the following, three benchmarking schemes are considered to evaluate the performance of the proposed adaptive batch-size control. 

\begin{itemize}  
    \item \textbf{Fixed Batch Size}:
    We consider a conventional approach that ignores device heterogeneity and employs a fixed and uniform batch size across all devices throughout training. Four typical local batch size settings are evaluated and marked by: $b_{k,n}= 16$ (16-fixed), $b_{k,n}= 32$ (32-fixed), $b_{k,n}= 64$ (64-fixed), and $b_{k,n}= 128$ (128-fixed).


    \item \textbf{Dynamic Batch Sizes Assisted FL (DBFL)~\cite{shi2022talk}:} 
    A state-of-the-art adaptive batch size strategy that exponentially increases uniform local batch size across rounds to exploit GPU parallelism and balance communication and computation time. The initial batch size is set to 16, and the  incremental factor is computed based on the optimization algorithm proposed in~\cite{shi2022talk}.
    Note that this scheme assumes a time-invariant communication latency and homogeneous devices, which limits its effectiveness in dynamic and heterogeneous environments.

    \item \textbf{Without Local Batch-Size Adaptation (No Local Adap.)}: 
       Assign a suboptimal global batch size $B^*_U$ uniformly to all $K$ devices, where $B^*_U$ balances the C$^2$ tradeoff via
       \begin{equation}
          B^*_U = \arg\min_B  \left\lceil \frac{\hat\alpha}{\epsilon - {\hat\beta}/{B}} \right\rceil  \left[ \max_{k \in \mathcal{K}} \left( \bar{T}_k^{\sf{cmm}} + \frac{ HW B}{f_k K} \right) \right]. 
       \end{equation}
        This scheme focuses solely on optimizing the global batch size to balance the C$^2$ tradeoff, while overlooking local batch-size adaptation to device heterogeneity.

\end{itemize}

\subsection{Computation-communication Tradeoff}

Fig.~\ref{fig:convergence_rounds_per_round_latency} shows the effects of global batch size on convergence round and per-round latency for both the MNIST and CIFAR-10 setups. It is observed that the convergence round decreases with the rise of global batch size due to the enhanced quality of stochastic gradient estimates. However, a larger batch size triggers a heavier computational load at the device side, leading to higher per-round latency. 
These opposite effects, both governed by the global batch size, motivate the investigation of C$^2$ tradeoff under the criterion of E2E latency.

Fig.~\ref{fig:practical_vs_theoretical_latency} illustrates the C$^2$ tradeoff by comparing empirical E2E latency measurements with the theoretical predictions of the proposed surrogate function in \eqref{eq:continuous_overall_E2E_latency}. Specifically, as the global batch size increases, the measured E2E latency exhibits a unimodal trend: it first decreases driven by smaller convergence round, and then increases as the per-round computational overhead dominates. This observed behavior validates the tradeoff between communication and computation in minimizing E2E latency for FL systems, as established in Remark~\ref{remark:Communication_Computation_Tradeoff}. Notably, the surrogate function captures this unimodal trend and its analytically derived optimal batch size (via \eqref{eq:B_optimal}, marked by the five-pointed star) aligns with the minimum latency observed in experiments, thereby validating the accuracy of the surrogate function.

\subsection{Performance of Adaptive Batch-Size Control}

The performance of the proposed adaptive batch-size control is evaluated in Figs.~\ref{fig:E2E latency vs test accuracy} and~\ref{fig:Accuracy of Different Batch Scheme vs. E2E latency}. 
In Fig.~\ref{fig:E2E latency vs test accuracy}, the convergence performance of proposed scheme and baselines varying with time is illustrated under MNIST and CIFAR-10 datasets.
Several observations can be drawn from both datasets. 
First, the proposed scheme yields a faster convergence than the counterpart without local batch-size adaptation. 
This benefits from addressing the device heterogeneity in computation and communication capabilities by minimizing per-round latency through Lemma~\ref{lemma:latency_equilibrium}.
Second, the scheme without local adaptation is seen to outperform other uniform allocation baselines, underscoring the effectiveness of selecting the optimal global batch size according to the round‐batch scaling law in Corollary~\ref{col:round_batch_constraint} to balance the C$^2$ tradeoff. 
Furthermore, a performance gain from DBFL is observed in the proposed scheme due to the stiff incremental factor in~\cite{shi2022talk} failing to cope with time-varying channels.
Finally, relative to the baseline without local adaptation, the proposed scheme reduces E2E latency by 37.5\% on CIFAR-10, surpassing the 26.7\% reduction observed on MNIST. This disparity arises because the greater parameter complexity of CIFAR-10 models exacerbates latency variance from device heterogeneity, while longer convergence processes accumulate optimization benefits across more rounds. Consequently, real-time batch-size adaptation yields greater relative latency savings on CIFAR-10. These results underscore the value of our adaptive scheme for efficient large-scale model training in heterogeneous environments.
Fig.~\ref{fig:Accuracy of Different Batch Scheme vs. E2E latency} further measures E2E latency of each scheme at different accuracy thresholds. 
Similar to the finding in Fig.~\ref{fig:E2E latency vs test accuracy}, the proposed scheme attains superiority over the benchmarks at all targets.
Moreover, the performance gap progressively widens with increasing accuracy thresholds due to amplified cumulative advantages from real-time batch allocation during extended training rounds.

\begin{figure}[t]
    \centering
    \subfigure[MNIST]{
        \includegraphics[width=0.48\columnwidth]{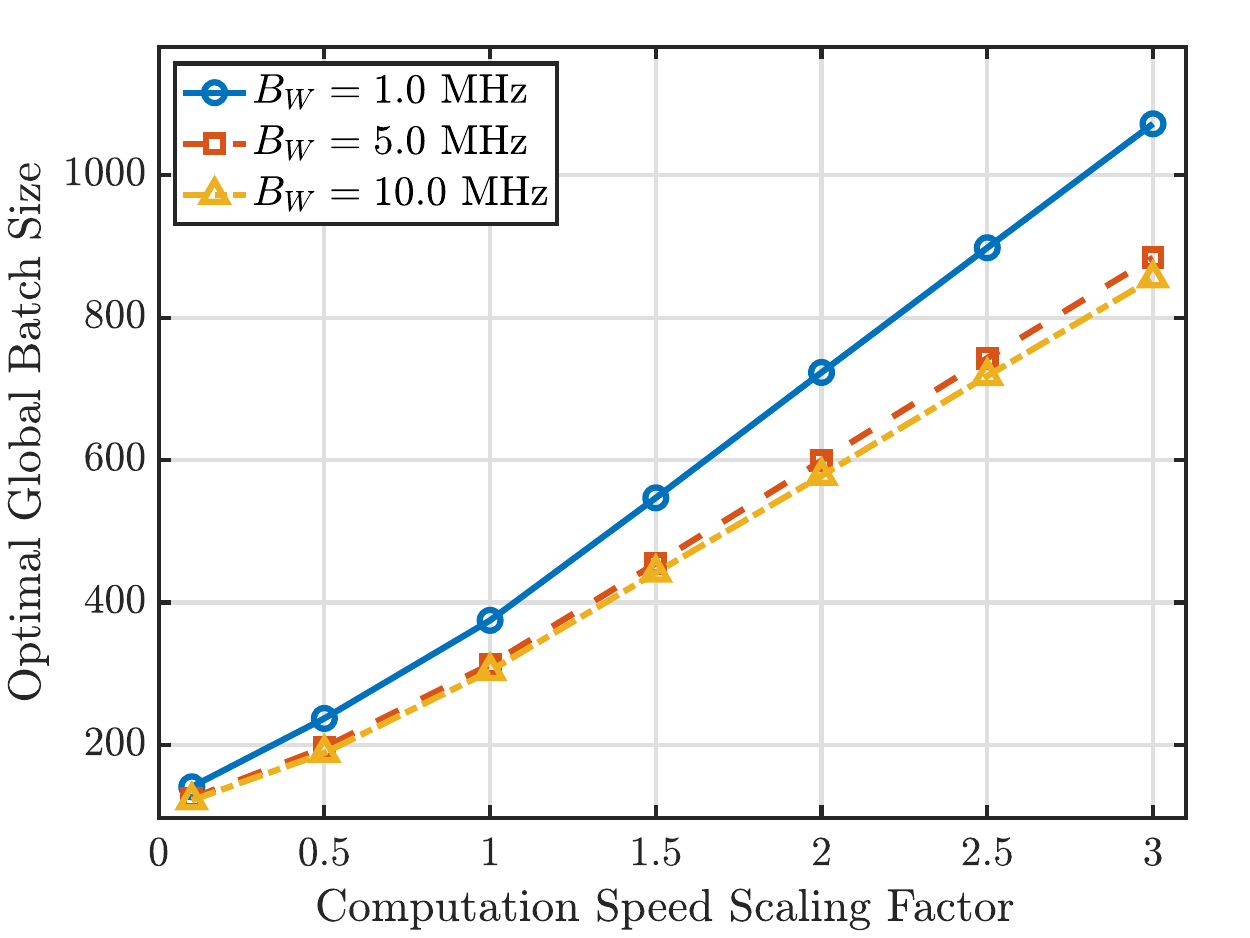}}
    \subfigure[CIFAR-10]{
        \includegraphics[width=0.48\columnwidth]{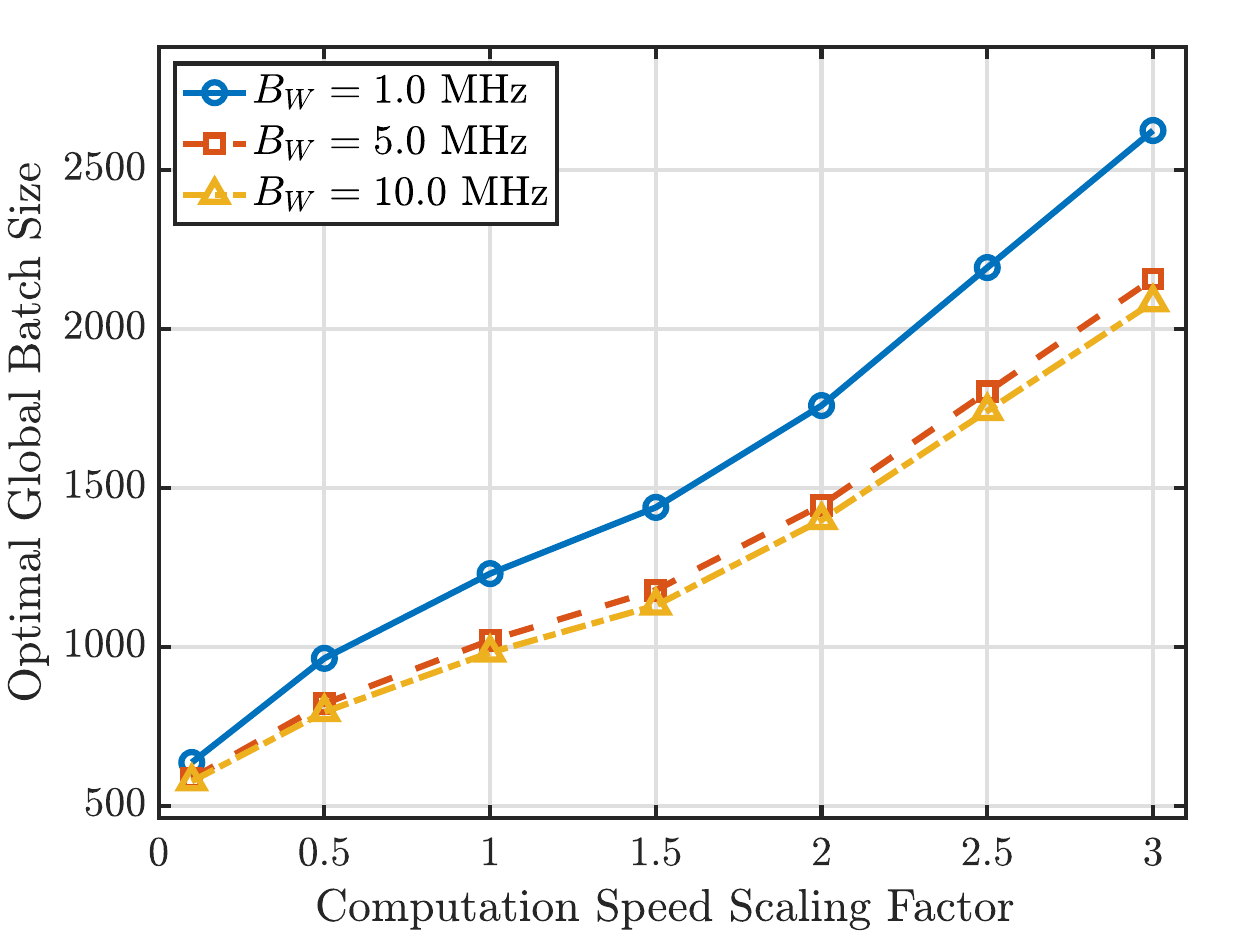}}

    \caption{Optimal global batch size vs. computation speed scaling factor (multiples of $\hat{f}$ to scale the system computation speed) at three system bandwidth levels.}
    \label{fig:Optimal Global batch size vs. Computation Speed Scaling Factor}
    \vspace{-3mm}
\end{figure}

\begin{figure}[t]
    \centering
    \subfigure[MNIST]{
        \includegraphics[width=0.48\columnwidth]{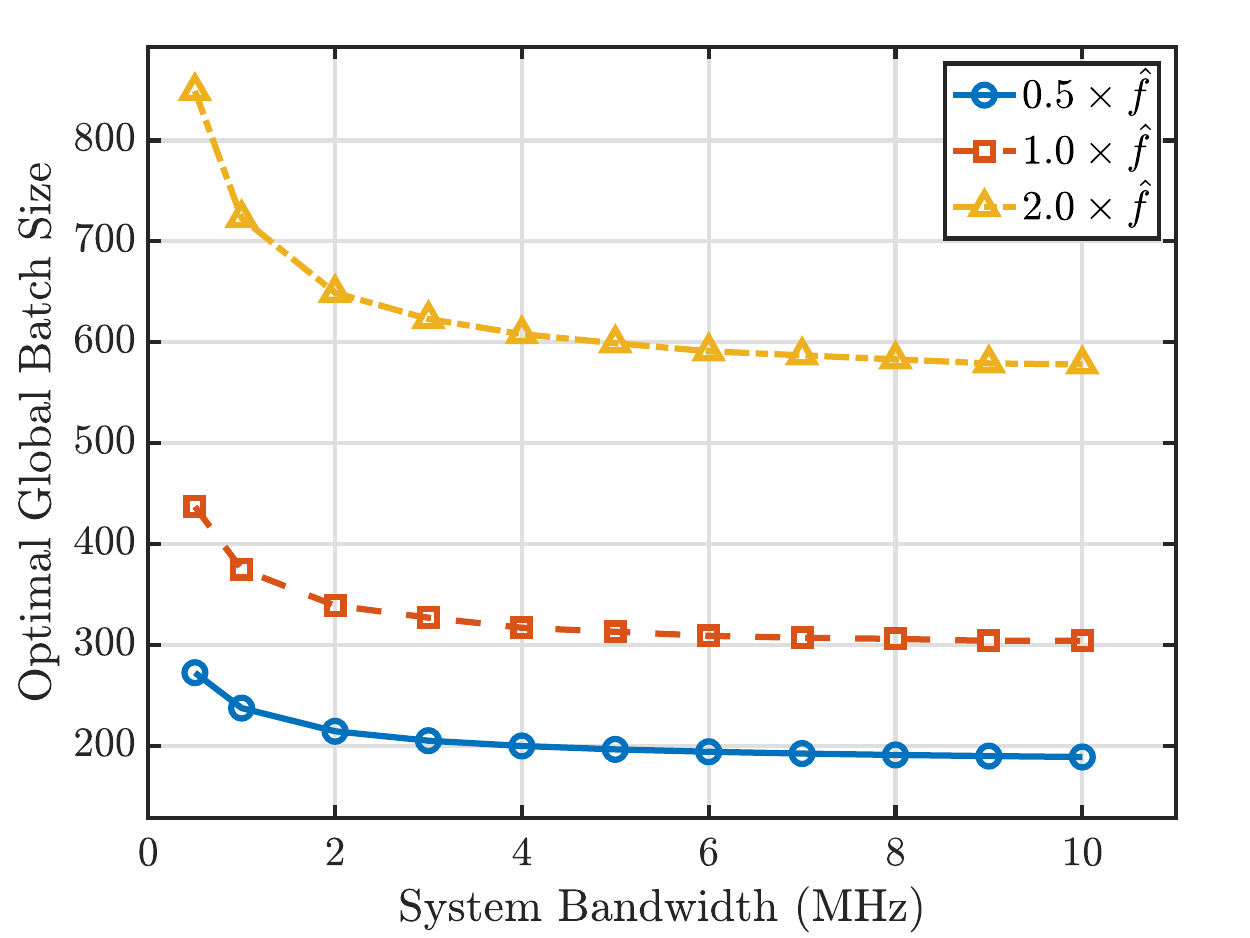}}
    \subfigure[CIFAR-10]{
        \includegraphics[width=0.48\columnwidth]{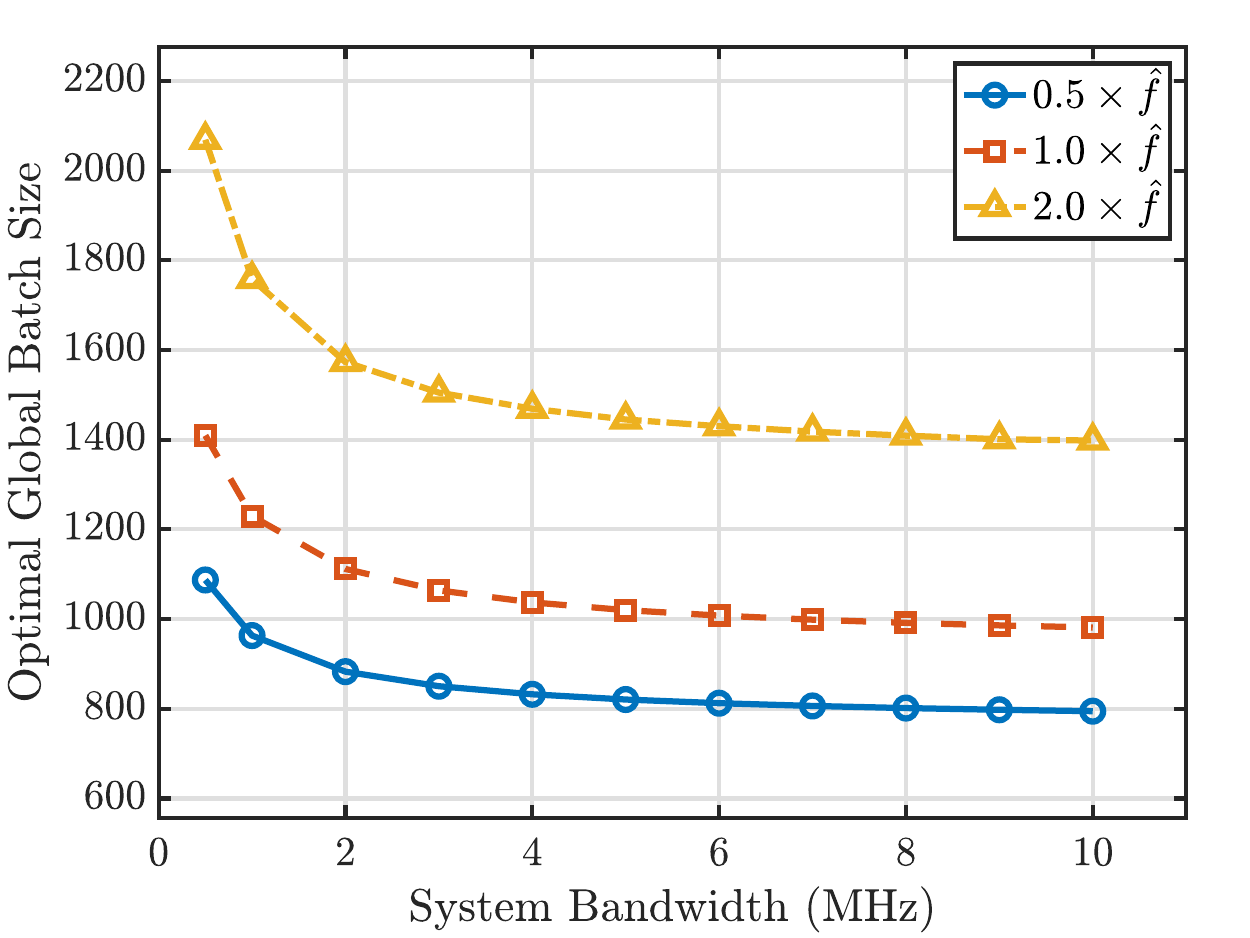}}

    \caption{Optimal global batch size vs. system bandwidth at three computation speed level, scaled by $\hat{f}$.}
    \label{fig:Optimal Global batch size vs. System Bandwidth}
    \vspace{-3mm}
\end{figure}

\subsection{Effects of System Parameters on Optimal Batch Size Configuration}

Fig.~\ref{fig:Optimal Global batch size vs. Computation Speed Scaling Factor} and~\ref{fig:Optimal Global batch size vs. System Bandwidth} illustrate the relation between the optimal global batch size and key system parameters, i.e., system bandwidth and computation speed, for both MNIST and CIFAR-10 datasets.
The results reveal two key insights arising from the C$^2$ tradeoff governed by global batch size. On the one hand,  Fig.~\ref{fig:Optimal Global batch size vs. Computation Speed Scaling Factor} demonstrates that under fixed bandwidth, elevated computational speed raises $B^*$, as enhanced device processing capability reduces the marginal cost of handling larger batches. This renders larger batches preferable by decreasing the convergence round without incurring prohibitive per-round computation latency. On the other hand, as shown in Fig.~\ref{fig:Optimal Global batch size vs. System Bandwidth}, at a fixed computation speed, increasing bandwidth reduces $B^*$. This is because higher bandwidth lowers communication latency, which shifts the bottleneck to computation time, making the per-round latency increase more rapidly with larger batch sizes.
In this context, a smaller global batch size is required to lower per-round latency at the cost of more rounds for convergence.

\section{Conclusion}
\label{sec:conclusion}



In this paper, we have proposed a batch-size control framework for low-latency FL implementations in heterogeneous device environments. Our convergence analysis characterizes the C$^2$ tradeoff between required communication rounds for satisfactory learning performance and per-round latency. 
Under slow‑fading channels, we balance the found tradeoff to minimize the E2E latency, yielding an optimal batch‑size control policy.
The framework is extended to fast-fading scenarios via real-time batch-size control, which is adaptive to device heterogeneity and instantaneous channel conditions.

This C$^2$ framework provides insightful guidelines for the efficient deployment of FL systems and suggests several promising research directions.
For instance, the C$^2$ tradeoff in FL systems can be extensively controlled by the key parameters of other techniques, such as gradient pruning, quantization, and model partitioning.
Moreover, characterizing the impact of these parameters on the performance of more advanced FL paradigms, including asynchronous updates and model splitting, remains an open research challenge.

\appendix

\subsection{Proof of Theorem~\ref{thm:H_avg_convergence}}\label{subsec:H_avg_proof}
The proof leverages fundamental properties of smooth optimization and stochastic gradient descent in FL systems. The global model update with $H>1$ is expressed as
\begin{equation}
\overline{\mathbf{w}}_{n+1} = \sum_{k=1}^{K}\frac{b_{k,n}}{B} \mathbf{w}_{k,n}^{(H)} = \sum_{k=1}^{K}\frac{b_{k,n}}{B} \left[ \overline{\mathbf{w}}_n - \eta\sum_{t=0}^{H-1} \mathbf{g}_{k,n}^{(t)} \right],
\end{equation}
According to Assumption \ref{AS:smoothness} and by taking the overall expectation on both sides, we obtain
\begin{equation}
\begin{aligned}
&\mathbb{E}\left[F(\overline{\mathbf{w}}_{n+1})\right] - \mathbb{E}\left[F(\overline{\mathbf{w}}_n)\right] \\
&\leq \mathbb{E}\left[\left\langle \nabla F(\overline{\mathbf{w}}_n), \overline{\mathbf{w}}_{n+1} - \overline{\mathbf{w}}_n \right\rangle \right]  + \frac{L}{2} \mathbb{E}\left[\| \overline{\mathbf{w}}_{n+1} - \overline{\mathbf{w}}_n \|^2 \right]\\
&\leq \underbrace{-\mathbb{E}\left[ \left\langle \nabla F(\overline{\mathbf{w}}_n), \eta \sum_{k=1}^{K}\frac{b_{k,n}}{B} \sum_{t=0}^{H-1} \mathbf{g}_{k,n}^{(t)} \right\rangle \right]}_{(a)} \\
& \quad + \underbrace{\frac{\eta^2L }{2}\mathbb{E}\left[ \left\| \sum_{k=1}^{K}\frac{b_{k,n}}{B} \sum_{t=0}^{H-1} \mathbf{g}_{k,n}^{(t)} \right\|^2\right]}_{(b)}.
\end{aligned}
\end{equation}
Next, we aim at finding the upper bounds for $(a)$ and $(b)$ respectively
\begin{equation}\label{eq:a}
\begin{aligned}
&(a) \overset{(a_1)}{\leq} -\eta \sum_{k=1}^{K}\frac{b_{k,n}}{B} \sum_{t=0}^{H-1} \mathbb{E}\left[ \left\langle \nabla F(\overline{\mathbf{w}}_n), \nabla F(\mathbf{w}_{k,n}^{(t)}) \right\rangle \right]\\
&\overset{(a_2)}{\leq} -\frac{\eta}{2} \sum_{k=1}^{K}\frac{b_{k,n}}{B} \sum_{t=0}^{H-1}\left( \mathbb{E}\left[\|\nabla F(\overline{\mathbf{w}}_n) \|^2 \right]  + \mathbb{E}\left[ \|\nabla F(\mathbf{w}_{k,n}^{(t)}) \|^2\right]\right. \\
&\qquad \left. - \mathbb{E}\left[ \|\nabla F(\overline{\mathbf{w}}_n)-\nabla F(\mathbf{w}_{k,n}^{(t)}) \|^2\right]\right)\\
&\overset{(a_3)}{\leq} - \frac{\eta(H+1)}{2} \mathbb{E}\left[\|\nabla F(\overline{\mathbf{w}}_n) \|^2 \right] 
   -\frac{\eta}{2} \sum_{k=1}^{K}\frac{b_{k,n}}{B} \sum_{t=1}^{H-1} \\
&\qquad \left( \mathbb{E}\left[ \|\nabla F(\mathbf{w}_{k,n}^{(t)}) \|^2\right] 
   + L^2 \underbrace{\mathbb{E}\left[ \|\overline{\mathbf{w}}_n-\mathbf{w}_{k,n}^{(t)} \|^2\right]}_{(c)} \right),
\end{aligned}
\end{equation}
where $(a_1)$ follows the linearity of expectation and the unbiasedness of Assumption \ref{AS:unbias_gradients}; $(a_2)$ applies the elementary inequality $2\langle z_1,z_2\rangle\geq\|z_1\|^2+\|z_2\|^2-\|z_1-z_2\|^2$; $(a_3)$ results from 
$\mathbf{w}_{k,n}^{(0)}=\overline{\mathbf{w}}_n$, when $t=0$, combined with Assumption~\ref{AS:smoothness}.
We further note that
\begin{equation}
\begin{aligned}
&(c)=\eta^2 \mathbb{E}\left[ \left\|\sum_{i=0}^{t-1}\mathbf{g}_{k,n}^{(i)}\right\|^2\right] \overset{(c_1)}{\leq} t\eta^2 \mathbb{E}\left[ \sum_{i=0}^{t-1}\|\mathbf{g}_{k,n}^{(i)}\|^2\right]\overset{(c_2)}{=}\\
&t\eta^2\left( \mathbb{E}\left[ \sum_{i=0}^{t-1}\|\mathbf{g}_{k,n}^{(i)}-\nabla F(\mathbf{w}_{k,n}^{(i)})\|^2\right]  + \mathbb{E}\left[ \sum_{i=0}^{t-1}\|\nabla F(\mathbf{w}_{k,n}^{(i)})\|^2\right]\right)\\
&\overset{(c_3)}{\leq} t^2 \eta^2\frac{\sigma^2}{b_{k,n}}+t\eta^2\mathbb{E}\left[ \sum_{i=0}^{t-1}\|\nabla F(\mathbf{w}_{k,n}^{(i)})\|^2\right],
\end{aligned}
\end{equation}
where $(c_1)$ follows from $\left\|\sum_{i=1}^{n} z_i\right\|^2 \leq n \sum_{i=1}^{n} \|z_i\|^2$ derived from the Cauchy-Schwarz inequality; $(c_2)$ applies the variance decomposition formula $\mathbb{E}\left[\|\mathbf{Z}\|^2\right] = \mathbb{E}\left[\|\mathbf{Z}-\mathbb{E}[\mathbf{Z}]\|^2\right] + \|\mathbb{E}[\mathbf{Z}]\|^2$; $(c_3)$ follows from Assumption \ref{AS:bounded_variance}.
\begin{equation}
\begin{aligned}
&\frac{\eta L^2}{2}\sum_{k=1}^{K}\frac{b_{k,n}}{B} \sum_{t=1}^{H-1}\mathbb{E}\left[ \|\overline{\mathbf{w}}_n-\mathbf{w}_{k,n}^{(t)} \|^2\right] \\
&\leq \frac{\eta L^2}{2}\sum_{k=1}^{K}\frac{b_{k,n}}{B} \sum_{t=1}^{H-1}\left[t^2 \eta^2\frac{\sigma^2}{b_{k,n}} +t\eta^2\mathbb{E}\left[ \sum_{i=0}^{t-1}\|\nabla F(\mathbf{w}_{k,n}^{(i)})\|^2\right]\right]\\
&\overset{(d_1)}{\leq} \frac{\eta^3 L^2 \sigma^2}{2} \sum_{k=1}^{K} \frac{b_{k,n}}{B} \frac{(2H-1)H(H-1)}{b_{k,n} 6}\\
&+\frac{\eta^3 L^2 }{2} \sum_{k=1}^{K} \frac{b_{k,n}}{B}\left[\sum_{i=0}^{H-2} \mathbb{E}\left[\|\nabla F(\mathbf{w}_{k,n}^{(i)})\|^2\right]\cdot \sum_{t=i+1}^{H-1}t \right]\\
&\leq \frac{\eta^3 L^2 \sigma^2 K (2H-1)H(H-1) }{12B}\\
& +\frac{\eta^3 L^2 }{2} \sum_{k=1}^{K} \frac{b_{k,n}}{B} \left(\sum_{i=0}^{H-1} \frac{(H-i-1)(H+i)}{2} \mathbb{E}\left[ \|\nabla F(\mathbf{w}_{k,n}^{(i)}) \|^2\right] \right)\\
&\leq \frac{\eta^3 L^2 \sigma^2 K (2H-1)H(H-1) }{12B}+\frac{\eta^3 L^2 H (H-1)}{4} \\
& \times \left(\mathbb{E}\left[\|\nabla F(\overline{\mathbf{w}}_n) \|^2 \right]+ \sum_{k=1}^{K} \frac{b_{k,n}}{B}\sum_{t=1}^{H-1}\mathbb{E}\left[ \|\nabla F(\mathbf{w}_{k,n}^{(t)}) \|^2\right] \right),
\end{aligned}
\end{equation}
where $(d_1)$ uses the sum of squares formula $\sum_{t=1}^{H-1} t^2 = \frac{(H-1)H(2H-1)}{6}$ and changes the order of summation.
We plug the above inequality back into \eqref{eq:a} and get
\begin{equation}
\begin{aligned}
&(a)
\leq -\frac{\eta}{2} \left( H+1 - \frac{ \eta^2 L^2 H(H-1)}{2} \right) \mathbb{E}\left[ \left\| \nabla F(\overline{\mathbf{w}}_n) \right\|^2  \right]\\
& -\frac{\eta}{2} \left( 1 - \frac{ \eta^2 L^2 H(H-1)}{2} \right)\sum_{k=1}^{K}\frac{b_{k,n}}{B} \sum_{t=1}^{H-1} \mathbb{E} \left[\| \nabla F(\mathbf{w}_{k,n}^{(t)}) \|^2 \right]\\
& +\frac{ \eta^3 L^2 \sigma^2 K(2H-1) H(H-1)}{12B}.
\end{aligned}
\end{equation}

To bound $(b)$, we apply similar analysis
\begin{equation}
\begin{aligned}
&(b) \leq \frac{\eta^2 L H}{2}\mathbb{E}\left[ \sum_{t=0}^{H-1}\left\| \sum_{k=1}^{K}\frac{b_{k,n}}{B}  \mathbf{g}_{k,n}^{(t)} \right\|^2\right]\\
&=\frac{ \eta^2 L H}{2}\mathbb{E}\left[ \sum_{t=0}^{H-1}\left\| \sum_{k=1}^{K}\frac{b_{k,n}}{B} \left( \mathbf{g}_{k,n}^{(t)} -\nabla F(\mathbf{w}_{k,n}^{(t)})\right)\right\|^2\right]\\
& \quad +\frac{\eta^2 L H}{2}\mathbb{E}\left[ \sum_{t=0}^{H-1}\left\| \sum_{k=1}^{K}\frac{b_{k,n}}{B} \nabla F(\mathbf{w}_{k,n}^{(t)})\right\|^2\right]\\
&\overset{(b_1)}\leq \frac{ \eta^2 L H^2}{2} \sum_{k=1}^{K}\left(\frac{b_{k,n}}{B}\right)^2\frac{\sigma^2}{b_{k,n}}\\
& \quad + \frac{\eta^2 L H}{2}\mathbb{E}\left[ \sum_{t=0}^{H-1}\left\| \sum_{k=1}^{K}\frac{b_{k,n}}{B} \nabla F(\mathbf{w}_{k,n}^{(t)})\right\|^2\right]\\
&\overset{(b_2)}{\leq} \frac{\eta^2 L H^2 \sigma^2}{2B}+\frac{\eta^2 L H}{2}\sum_{k=1}^{K}\frac{b_{k,n}}{B} \sum_{t=0}^{H-1} \mathbb{E}\left[ \left\|  \nabla F(\mathbf{w}_{k,n}^{(t)})\right\|^2\right],
\end{aligned}
\end{equation}
where $(b_1)$ follows because each $\mathbf{g}_{k,n}^{(t)} -\nabla F(\mathbf{w}_{k,n}^{(t)})$ has zero mean and is independent, and from Assumption~\ref{AS:bounded_variance}; $(b_2)$ applies Jensen’s inequality $\left\|\sum_{k=1}^{K}\frac{b_{k}}{B}z_k\right\|^2 \leq \sum_{k=1}^{K}\frac{b_{k}}{B}\left\|z_k\right\|^2$.

Combining above results, we have
\begin{equation}
\begin{aligned}
&\mathbb{E}\left[ F(\overline{\mathbf{w}}_{n+1}) \right] - \mathbb{E}\left[F(\overline{\mathbf{w}}_n) \right]\\
&\leq -\frac{\eta}{2} \left(H+ 1 - \frac{\eta^2 L^2 H(H-1)}{2} - \eta L H \right) \mathbb{E}\left[ \left\| \nabla F(\overline{\mathbf{w}}_n) \right\|^2  \right]\\
&\quad -\frac{\eta}{2} \left( 1 - \frac{\eta^2 L^2H(H-1)}{2} - \eta L H \right)\\
& \quad \times \sum_{k=1}^{K}\frac{b_{k,n}}{B}\sum_{t=1}^{H-1} \mathbb{E} \left\| \nabla F(\mathbf{w}_{k,n}^{(t)}) \right\|^2 + \frac{\eta^2 L H^2 \sigma^2}{2B}\\
&\quad + \frac{\eta^3 L^2 \sigma^2 K (2H-1)H(H-1)}{12B} .
\end{aligned}
\end{equation}

Under the condition \( 1 \geq \frac{L^2\eta^2 H(H-1)}{2} + \eta L H \), the second term on the RHS can be discarded
\begin{equation}
\begin{aligned}
&\mathbb{E}\left[ F(\overline{\mathbf{w}}_{n+1}) \right] - \mathbb{E}\left[F(\overline{\mathbf{w}}_n) \right]\leq -\frac{\eta H}{2}  \mathbb{E}\left[ \left\| \nabla F(\overline{\mathbf{w}}_n) \right\|^2  \right]\\
&\quad+ \frac{\eta^2 L H \sigma^2 }{2B} \left(H+\frac{\eta L K (2H-1)(H-1)}{6} \right).
\end{aligned}
\end{equation}

Dividing both sides by $\frac{\eta H} {2}$ and rearranging terms yield
\begin{equation}
\begin{aligned}
&\mathbb{E}\left[ \left\| \nabla F(\overline{\mathbf{w}}_n) \right\|^2  \right] \leq \frac{2 \left(\mathbb{E}\left[ F(\overline{\mathbf{w}}_n) \right] - \mathbb{E}\left[F(\overline{\mathbf{w}}_{n+1}) \right] \right)}{\eta H} \\
&\quad + \frac{\eta L \sigma^2 }{B} \left(H+\frac{\eta L K (2H-1)(H-1)}{6} \right).
\end{aligned}
\end{equation}

By summing over $n \in \{0, \ldots, N-1\}$  and dividing by $N$, we obtain 
\begin{equation}
\begin{aligned}
&\frac{1}{N}\sum_{n=0}^{N-1} \mathbb{E}\left[ \left\| \nabla F(\overline{\mathbf{w}}_n) \right\|^2  \right] \leq \frac{2 \left( F(\overline{\mathbf{w}}_0)  - \mathbb{E}\left[F(\overline{\mathbf{w}}_N) \right] \right)}{N\eta H}\\
&+ \frac{\eta L \sigma^2 }{B} \left(H+\frac{\eta L K (2H-1)(H-1)}{6} \right)\\
&\leq \frac{2 \left( F(\overline{\mathbf{w}}_0)  - F^*\right)}{N\eta H} + \frac{\eta L \sigma^2 }{B} \left(H+\frac{\eta L K (2H-1)(H-1)}{6} \right),
\end{aligned}
\end{equation}
where the second inequality follows Assumption~\ref{AS:Bounded Loss Function}.

The proof for \(H = 1\) follows an identical logical flow to  above \(H > 1\) case. However, when \(H = 1\), terms involving client drift (e.g., $(c)$ in \eqref{eq:a}) vanish since \(\mathbf{w}_{k,n}^{(t)} \equiv \overline{\mathbf{w}}_n\) for all clients and multi-step summations collapse. For conciseness, we omit independent proof for \(H = 1\), as the specialized result is subsumed by Theorem~\ref{thm:H_avg_convergence}. This completes the proof.

\subsection{Proof of Lemma~\ref{lemma:latency_equilibrium}}
\label{subsec:latency_balance_proof}

We establish the latency equilibrium principle through contradiction. Consider the system of two arbitrary devices \(k_1, k_2  \) under optimal allocation \(\{\tilde{b}_k^\star\}\) with per-device latencies $\tau_{k_i} \triangleq T^{\sf{cmm}}_{k_i} + \frac{HW \tilde{b}_{k_i}^\star}{f_{k_i}},  i \in \{1,2\}$. Without loss of generality, assume \(\tau_{k_1} > \tau_{k_2}\), implying the per-round latency \(\tau_{\min}(B) = \tau_{k_1}\).
Construct a perturbed allocation by transferring \(\delta > 0\) batches from \(k_1\) to \(k_2\): $\tilde{b}_{k_1}' = \tilde{b}_{k_1}^\star - \delta, \;\tilde{b}_{k_2}' = \tilde{b}_{k_2}^\star + \delta$, yielding new latencies $\tau_{k_1}' = \tau_{k_1} - \frac{HW\delta}{f_{k_1}} ,\;
\tau_{k_2}' = \tau_{k_2} + \frac{HW\delta}{f_{k_2}}$.
The latency difference becomes $\tau_{k_1}' - \tau_{k_2}' = (\tau_{k_1} - \tau_{k_2}) - HW\delta\left(\frac{1}{f_{k_1}} + \frac{1}{f_{k_2}}\right)$.

Select \(\delta = \frac{f_{k_1}f_{k_2}(\tau_{k_1} - \tau_{k_2})}{HW(f_{k_1} + f_{k_2})}>0\), which yields
$\tau_{k_1}' = \tau_{k_2}' = \tau_{k_1} - \frac{W\delta}{f_{k_1}} < \tau_{k_1}$.
This new latency contradicts the optimality of \(\tau_{\min}(B)\). Similar reasoning applies for \(\tau_{k_1} < \tau_{k_2}\). By induction, latency equivalence holds across all devices \(k\).

\subsection{Proof of Theorem~\ref{thm:optimal_coordination}}
\label{subsec:optimal_batch_proof}

We begin by analyzing the surrogate function $\tilde{\psi}(B)$ and proving its unimodality, from which the optimal global batch size in~\eqref{eq:B_optimal} is derived.

From~\eqref{eq:continuous_overall_E2E_latency}, for $B \in \left( \frac{\hat{\beta}}{\epsilon},\, \tilde{B}^{\sf th} \right]$, the derivative of $\tilde{\psi}(B)$ is 
\begin{equation}
    \tilde\psi'(B)
   = -\,\hat\alpha\,\frac{\hat\beta\,\tau_{1b}}{(\epsilon B-\hat\beta)^2}
   <0,
\end{equation}indicating that $\tilde{\psi}(B)$ is strictly decreasing over this interval.

For $B > \tilde{B}^{\sf th}$, the derivative becomes
 \begin{equation}
     \tilde\psi'(B)
   = \frac{\hat\alpha (W\epsilon B^2- 2W\hat\beta B- \hat f_\Sigma\hat\beta)}{f_\Sigma\,(\epsilon B-\hat\beta)^2}.
 \end{equation}
The numerator \(p(B) = W\epsilon B^2 - 2W\hat{\beta} B - \hat{f}_{\Sigma} \hat{\beta}\) is a convex quadratic function that opens upwards and has a unique positive zero, given by 
\begin{equation}
    B_{\epsilon} = \frac{\hat{\beta}}{\epsilon} \left(1 +\sqrt{1 + \frac{\hat{f}_{\Sigma} \epsilon}{W\hat{\beta}}}\right).
\end{equation}
Thus, $\tilde{\psi}(B)$ is strictly decreasing on $(\tilde{B}^{\sf th},\, B_{\epsilon})$ and strictly increasing on $(B_{\epsilon},\, B_{\max}]$, implying a unique minimizer at $B_{\epsilon}$. 

The definition of $\tau_{1b}$ directly implies continuity of $\tilde{\psi}(B)$ at $\tilde{B}^{\sf th}$. Thus, combining both regions, $\tilde{\psi}(B)$ is unimodal over $\left( \frac{\hat{\beta}}{\epsilon}, B_{\max} \right]$, with the global minimizer given by \begin{equation}
    \tilde{B}^* = \max\left( \tilde{B}^{\sf th}, B_{\epsilon} \right) .
\end{equation}

For integer constraint $B \in \mathbb{Z}^+$, we set $\tilde{B}^*$ to the integer according to the surrogate evaluation, yielding the result in~\eqref{eq:B_optimal}.
Finally, the $\max$ operator in~\eqref{eq:B_optimal} ensures $B^* \ge B^{\sf th}$, thereby satisfying the latency equilibrium principle in Lemma~\ref{lemma:latency_equilibrium}. Substituting $B^*$ into the allocation formula~\eqref{eq:allocation_principle} directly yields the optimal device-level batch sizes as given in~\eqref{eq:device_allocation}.

\bibliography{main}
\bibliographystyle{IEEEtran}

\end{document}